\documentclass[letterpaper]{article} 
\usepackage{aaai23}  
\usepackage{times}  
\usepackage{helvet}  
\usepackage{courier}  
\usepackage[hyphens]{url}  
\usepackage{graphicx} 
\usepackage{natbib}  
\usepackage{caption} 
\usepackage{algorithm}
\usepackage{algorithmic}
\usepackage{multirow}
\usepackage{newfloat}
\usepackage{listings}
\DeclareCaptionStyle{ruled}{labelfont=normalfont,labelsep=colon,strut=off} 
\floatstyle{ruled}
\newfloat{listing}{tb}{lst}{}
\floatname{listing}{Listing}
\title{Rethinking Disparity: A Depth Range Free Multi-View Stereo Based on Disparity}
\author{
    Qingsong Yan \equalcontrib \textsuperscript{,\rm 1,\rm 2}, 
    Qiang Wang \equalcontrib \textsuperscript{,\rm 3}, 
    Kaiyong Zhao \textsuperscript{\rm 4}, 
    Bo Li \textsuperscript{\rm 2}, 
    Xiaowen Chu \thanks{Corresponding author} \textsuperscript{,\rm 2,5}, 
    Fei Deng \textsuperscript{\rm \textdagger,\rm 1}  
}
\affiliations{
	\textsuperscript{1} Wuhan University, Wuhan, China \\
	\textsuperscript{2} The Hong Kong University of Science and Technology, Hong Kong SAR, China\\
	\textsuperscript{3} Harbin Institute of Technology (Shenzhen), Shenzhen, China\\
	\textsuperscript{4} XGRIDS, Shenzhen, China \\
	\textsuperscript{5} The Hong Kong University of Science and Technology (Guangzhou), Guangzhou, China\\
	{\tt\small  yanqs\_whu@whu.edu.cn, qiang.wang@hit.edu.cn, kyzhao@xgrids.com, bli@cse.ust.hk } \\
	{\tt\small xwchu@ust.hk, fdeng@sgg.whu.edu.cn }

}

\usepackage{bibentry}

\begin{document}

	\maketitle
	
	\begin{abstract}
		
		Existing learning-based multi-view stereo (MVS) methods rely on the depth range to build the 3D cost volume and may fail when the range is too large or unreliable. To address this problem, we propose a disparity-based MVS method based on the epipolar disparity flow (E-flow), called DispMVS, which infers the depth information from the pixel movement between two views. The core of DispMVS is to construct a 2D cost volume on the image plane along the epipolar line between each pair (between the reference image and several source images) for pixel matching and fuse uncountable depths triangulated from each pair by multi-view geometry to ensure multi-view consistency. To be robust, DispMVS starts from a randomly initialized depth map and iteratively refines the depth map with the help of the coarse-to-fine strategy. Experiments on DTUMVS and Tanks\&Temple datasets show that DispMVS is not sensitive to the depth range and achieves state-of-the-art results with lower GPU memory.
		
	\end{abstract}

	\section{Introduction}
	
	Multi-view stereo matching (MVS) is a core technique in 3D reconstruction that has been extensively studied \cite{goesele2007multi,furukawa2009accurate,galliani2015massively,schoenberger2016mvs}. Although traditional methods try to introduce additional constraints \cite{xu2019multi,romanoni2019tapa,xu2020marmvs,wang2020mesh,xu2020planar} to deal with textureless regions or repeated textures,  they still have difficulty in guaranteeing the generation of high-quality point clouds in many cases.
	
	\begin{figure}[t]
		
		\begin{center}
			\includegraphics[width=1.0\linewidth]{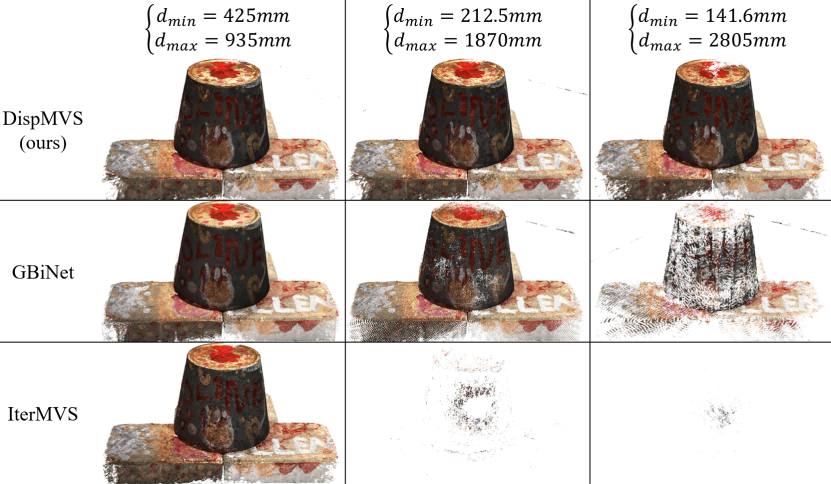}
		\end{center}
		
		\caption{
			\textbf{The influence of the depth range.} This figure compares DispMVS with two state-of-the-art methods ( GBiNet \cite{mi2022generalized} and IterMVS \cite{wang2022itermvs}  ) and shows that DispMVS can still generate high-quality point clouds when changing the depth range by Eq. \ref{eq:change_depth_range}. Unlike other methods that use the depth range to build a 3D cost volume and estimate the depth map, DispMVS only builds a 2D cost volume along the epipolar line and uses the multi-view geometry to estimate the depth map.	}
		
		\label{fig:dtu_range}
		\vspace{-1em}
		
	\end{figure}

	Recently, learning-based methods have brought a new light to MVS. MVSNet \cite{yao2018mvsnet} shows a fully differentiable pipeline, which firstly uses a convolutional neural network (CNN) to extract features from input images,and then splits the 3D space into several bins covering a certain depth range to build a 3D cost volume by differentiable homograph, and finally relies on a 3D CNN to regress the depth map. Although MVSNet achieves impressive results on several public benchmarks \cite{aanaes2016large,knapitsch2017tanks}, it is not efficient and requires a lot of GPU memory to infer. Therefore, the coarse-to-fine strategy \cite{gu2020cascade,yang2020cost,cheng2020deep,mi2022generalized} and the recurrent network \cite{yao2019recurrent,wei2021aa,wang2021patchmatchnet,wang2022itermvs} are used to upgrade the MVSNet. The coarse-to-fine strategy uses the coarse stage to recover a coarse depth map and reduces the number of bins needed at the fine stage to reduce the GPU memory. Furthermore, the recurrent network, such as LSTM \cite{shi2015convolutional} and GRU \cite{cho2014learning}, can directly replace the 3D CNN to process the 3D cost volume with a lower GPU memory footprint during the inference. Since these strategies may sacrifice accuracy due to the reduced search range and the lack of global context, many studies have improved the design of loss functions and feature processing modules. In terms of loss functions, there is not only the regression-based method to guide continuous depth \cite{yao2018mvsnet,gu2020cascade} but also the classification-based method to guide discrete depth through 3D cost volume \cite{yao2019recurrent,mi2022generalized}, or a mixture of them to achieve better depth prediction quality \cite{peng2022rethinking}. As for the feature processing module, the deformation convolution \cite{wei2021aa}, the attention mechanism \cite{luo2020attention}, and the Transformer \cite{zhu2021multi} are used to improve the quality of image features. In addition, the visibility information \cite{xu2020pvsnet,zhang2020visibility} and the epipolar \cite{ma2021epp,yang2022mvs2d} are also used to boost the performance.

	A critical issue of those existing methods is the fixed depth range in building the cost volume \cite{cheng2020deep,mi2022generalized}. Usually, the depth range decides the 3D distribution of cost volume that the network attempts to fit, and the size of cost volume is limited to the computational and memory capability, which results in these methods can be easy to over-fit the configured depth range. Fig. \ref{fig:dtu_range} shows that the quality of point cloud construction by two state-of-the-art methods, GBiNet \cite{mi2022generalized} and IterMVS \cite{wang2022itermvs}, is dramatically degraded when the depth range is enlarged. The reason is that these methods cannot capture enough matching information with a fixed number of depth bins.
	
	In this paper, we propose a new MVS pipeline, which allows the CNN to focus only on the matching problem between two different views and relies on the multi-view geometry to recover the depth by matching results. The contributions of this paper are as follows.
	
	\begin{itemize}
		
		\item Instead of constructing the 3D cost volume, this paper only constructs the 2D cost volume to match pixels between each pair and generates the depth map by triangulation. In other words, DispMVS exploits the multi-view geometry to reduce the burden of networks, which does not rely on the computationally expensive 3D cost volume to estimate the depth.
		\item We redesign the flow to deal with the multi-view stereo matching without applying stereo rectification for each pair. First, we propose the epipolar disparity flow (E-flow) and reveal the relationship between the E-flow and the depth. Then we extend E-flow from two-view to multi-view and iteratively update the E-flow by a fused depth to maintain multi-view consistency.
		\item DispMVS achieves the state-of-the-art result on the DTUMVS and Tanks\&Temple without the 3D cost volume, demonstrating the effectiveness of combining multi-view geometry with a learning-based approach.
		
	\end{itemize}

	\section{Related Work}
	
	With decades of development of MVS, many traditional and learning methods are proposed. Traditional MVS cannot surpass the limitations of the artificially designed matching pipeline and fail to reconstruct the non-Lambertian regions. On the contrary, learning-based methods can automatically find the most helpful information in a data-driven manner, and the benchmarking results of \cite{aanaes2016large,knapitsch2017tanks} show that the learning-based method can easily outperform traditional methods. Generally, the learning-based methods build a 3D cost volume, and we can categorize them into 3D convolution-based and RNN-based methods according to how they handle the 3D cost volume.
	
	\paragraph{Traditional MVS}
	Traditional MVS has three major types: volumetric method, point cloud method, and depth map method. The volumetric method \cite{seitz1999photorealistic,kutulakos2000theory,kostrikov2014probabilistic} splits the space into inside and outside and cuts out the surface. The point cloud method \cite{lhuillier2005quasi,goesele2007multi,furukawa2009accurate} reconstructs dense point clouds from sparse point clouds. Although these methods can reconstruct high-quality results, the volumetric and point cloud methods require lots of GPU memory and are hard to parallelize. The depth map method is the most popular, which separately reconstructs the depth map of each view and fuses them to generate the point cloud \cite{merrell2007real,galliani2015massively}. Shen \cite{shen2013accurate} and Colmap \cite{zheng2014patchmatch,schoenberger2016mvs} extends the PatchMatch \cite{bleyer2011patchmatch} to multi-view and simultaneously estimates the  normal and the depth. Meanwhile, Gipuma \cite{galliani2015massively} and ACMM \cite{xu2019multi} use a GPU to improve the computational efficiency. The superpixel  \cite{romanoni2019tapa}, the plane \cite{xu2020planar}, and the mesh \cite{wang2020mesh} are used to reduce mismatching in non-Lambertian regions. Although these methods can achieve stable results, they cannot surpass the limitation of hand-craft methods in a challenging environment.

	\paragraph{3D Convolution Method}
	
	Various approaches have been proposed to address the shortcomings of MVSNet \cite{yao2018mvsnet}.
	
	The straightforward solution to reduce the GPU memory is building fewer 3D cost volume. Fast-MVSNet \cite{yu2020fast} only calculates the 3D cost volume on a sparse depth map and propagates the sparse depth map into a dense depth map. CasMVSNet \cite{gu2020cascade} and CVP-MVSNet \cite{yang2020cost} uses a coarse-to-fine strategy to deal with the 3D cost volume and reduce computation cost on the high resolution. GBiNet \cite{mi2022generalized} treats the MVS as a binary search problem and only builds the 3D cost volumes on the side with a high probability of containing the depth.
	
	Various methods are proposed to improve the 3D cost volume quality. P-MVSNet \cite{luo2019p} uses the isotropic and anisotropic 3D convolution to estimate the depth map. AttMVS \cite{luo2020attention} applies attention mechanism into the 3D cost volume to improve the robustness. UCS-MVSNet \cite{cheng2020deep} uses the uncertainty as a guide to adjust the 3D cost volume. EPP-MVSNet \cite{ma2021epp} propose an epipolar-assembling module to enhance the 3D cost volume. MVSTR \cite{zhu2021multi} relies on the 3D-geometry transformer \cite{dosovitskiy2020image} to obtain global context and 3D consistency. Also, considering the occlusion, Vis-MVSNet  \cite{zhang2020visibility} and PVSNet \cite{xu2020pvsnet} introduce the visibility to filter out unreliable regions. Besides, MVSCRF \cite{xue2019mvscrf} uses a conditional random field to ensure the smoothness of the depth map, and Uni-MVSNet \cite{peng2022rethinking} combines the regression and classification by the unified focal loss.
	
	\begin{figure*}[t]
		
		\begin{center}
			\includegraphics[width=0.94\linewidth]{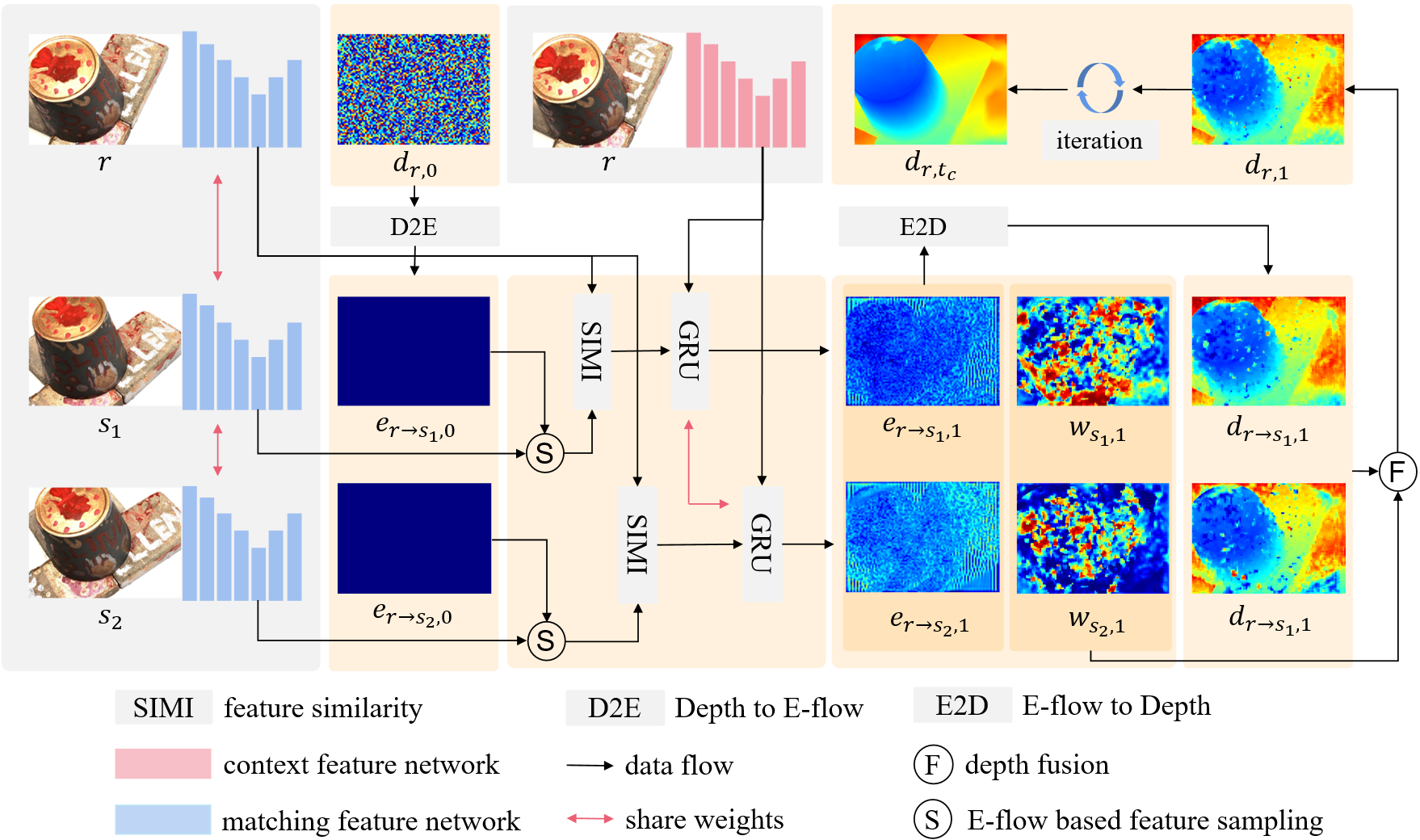}
		\end{center}
		
		\caption{
			\textbf{The pipeline of DispMVS.} After extracting features from input images, DispMVS first uses a random depth map to initialize the E-flow between each pair and then triangulates a new depth map by the E-flow that is updated through a GRU module. Finally, with several iterations, DispMVS can reconstruct a high-quality depth map. 
		}
		
		\label{fig:pipeline}
		\vspace{-1em}
	\end{figure*}
	
	In addition, unsupervised MVS has also achieved impressive results. 
	Un-MVSNet \cite{khot2019learning} uses photometric as a guide to learning the depth map. MVSNet2 \cite{dai2019mvs2} uses multi-view depth maps to filter out occlusion. M3VSNet \cite{huang2021m3vsnet} uses the normal to ensure smoothness and improve the depth map. JADAS \cite{xu2021self} combines depth and semantics to improve the depth map. RC-MVSNet \cite{chang2022rc} trains a NeRF \cite{mildenhall2020nerf} to maintain rendering consistency to solve ambiguity correspondences among views.

	\paragraph{RNN Method}
	
	Instead of using the 3D CNN to process the 3D cost volume, the RNN-based method uses the more efficient LSTM \cite{shi2015convolutional} or GRU \cite{cho2014learning}. R-MVSNet \cite{yao2019recurrent} uses the GRU to regularize the cost volume sequentially. RED-Net \cite{liu2020novel} utilizes a 2D recurrent encoder-decoder structure to process the cost volume based on the GRU module. D2HC-RMVSNet \cite{yan2020dense} and AA-RMVSNet  \cite{wei2021aa} proposes a hybrid architecture that combines the LSTM and the UNet \cite{ronneberger2015u}.
	PatchMatchNet \cite{wang2021patchmatchnet} introduces an end-to-end PatchMatch \cite{bleyer2011patchmatch} with adaptive propagation during each iteration and achieves competitive performance with lower GPU memory. Recently, IterMVS \cite{wang2022itermvs} uses RAFT \cite{teed2020raft,lipson2021raft} as backbone and iterative updates the depth map based on the GRU module.
	

	\section{Method}
	
	In this section, we introduce the details of the proposed method. The pipeline of DispMVS is demonstrated in Fig. \ref{fig:pipeline}. Unlike other methods depending on the depth range and differentiable homograph warping to build the 3D cost volume, we use a network to match pixels along the epipolar line and triangulate the depth. Therefore, we first discuss the relationship between the flow, the depth, and the E-flow. Then, we extend the E-flow to multi-view and explain the details of DispMVS and the loss function.

	\subsection{Flow and Depth}
	
	Given a reference view $r$ and a source view $s$ with their interior matrix $K_r,K_s$ and their relative exterior matrix $[R_s,T_s]$, we define $d_r,d_s$ as the depth, and $\vec{f}_{r \to s},\vec{f}_{s \to r}$ as the flow of each view. Assuming that the scene is static, we can convert depth to flow and vice versa according to the multi-view geometry \cite{hartley2003multiple}.
	
	\textbf{Depth} The depth describes the 3D shape of an image and can re-project a pixel on the image plane to 3D space. Eq. \ref{eq:re-project} re-projects a pixel $p_r$ in $r$ to $P_{p_r}$ in 3D by its depth $d_r(p_r)$, in which $\tilde{p}_r$ is the homogeneous representation of $p_r$ for computation efficiency. $P_{p_r}$ can also be projected to $s$ by Eq. \ref{eq:project}.
	
	\begin{eqnarray} 
		P_{p_r} &&= d_r(p_r)K_r^{-1}\tilde{p}_r \label{eq:re-project} \\
		p_s && \simeq K_s(R_s P_{p_r} + T_s)  \label{eq:project}
	\end{eqnarray}
	
	\textbf{Flow} The flow describes the movement of pixels on the image plane between two images. For a matched pixel pair $p_r$ in $r$ and $p_s$ in $s$, we calculate the flow $\vec{f}_{p_{r \to s}}$ by Eq. \ref{eq:flow}. Generally, the flow does not need to follow geometry constraints and has two degrees of freedom.
	
	\begin{eqnarray} 
		\vec{f}_{r \to s}(p_r) = p_s - p_r \label{eq:flow}
	\end{eqnarray}
	
	\textbf{Depth to Flow} Eq. \ref{eq:depth_to_flow} shows how to convert the the $d_r(p_r)$ to the $f_r(p_r)$, where $\Rightarrow$ denotes the conversion. We first re-project $p_r$ to $P_{p_r}$ by $d_r(p_r)$ as Eq. \ref{eq:re-project} shows and then project $P_{p_r}$ to $s$ by Eq. \ref{eq:project} to get the matched pixel $p_s$. Finally, we can calculate $\vec{f}_{r \to s}(p_r)$ by Eq. \ref{eq:flow}.
	
	\begin{eqnarray} 
		\label{eq:depth_to_flow}
		d_r(p_r) \Rightarrow  P_{p_r} \Rightarrow p_s \Rightarrow  \vec{f}_{r \to s}(p_r)
	\end{eqnarray}
	
	\textbf{Flow to Depth} Although triangulation is s straightforward method to convert $f_r$ to $d_r$ \cite{hartley2003multiple}, it has to solve a not differentiable homogeneous linear function. Considering this, we use a differentiable closed-form solution to calculate the depth, even though it is not optimal. Given $p_r$ and $f_{r \to s}(p_r)$, we can determine $p_s$ by Eq. \ref{eq:flow}. Based on multi-view geometric consistency, we have the constrain in Eq. \ref{eq:multi_depth_constrain}:
	
	\begin{eqnarray}
		\label{eq:multi_depth_constrain}
		d_r(p_r) K_r^{-1}\tilde{p}_r=R_s d_s(p_s) K_s^{-1}\tilde{p}_s+T_s
	\end{eqnarray}
	
	Let $T_s=(t_{sx},t_{sy},t_{sz})^T$, $K_r^{-1}\tilde{p}_r=(p_{rx},p_{ry},p_{rz})^T$ and $R_s K_s^{-1}\tilde{p}_s = (p_{sx},p_{sy},p_{sz})^T$. We can calculate $d_r(p_r)$ by Eq. \ref{eq:flow_to_depth}:
	
	\begin{eqnarray}
		\label{eq:flow_to_depth}
		\left\{\begin{array}{@{}l@{\quad}l}
			d_{xr}(p_r) &= (t_{sx}p_{sz}-t_{sz}p_{rx})/(p_{sx}p_{sz}-p_{sz}p_{rx}) \\[\jot]
			d_{yr}(p_r) &= (t_{sy}p_{sz}-t_{sz}p_{ry})/(p_{sy}p_{sz}-p_{sz}p_{rx}) 
		\end{array}\right.
	\end{eqnarray}
	
	Eq. \ref{eq:flow_to_depth} shows that there are two ways to compute the depth, namely $d_{xr}(p_r)$ and $d_{yr}(p_r)$, since $\vec{f}_{r\to s}(p_r)$ is a 2D vector that provides flow in $x$ dimension $\vec{f}_{xr\to xs}(p_r)$ and $y$ dimension $\vec{f}_{yr\to ys}(p_r)$. Theoretically, $d_{xr}(p_r)$ equals $d_{yr}(p_r)$. However, a smaller flow is not numerically stable and will bring noise into the triangulation. Therefore we select the depth triangulated by the larger flow by Eq. \ref{eq:select_depth}:
	
	\begin{eqnarray}
		\label{eq:select_depth}
		d_r(p_r) = \left\{\begin{array}{@{}l@{\quad}l}
			d_{xr}(p_r) & \mbox{if $|\vec{f}_{xr\to xs}(p_r)| \ge |\vec{f}_{yr\to ys}(p_r)|$} \\[\jot]
			d_{yr}(p_r) & \mbox{if $|\vec{f}_{xr\to xs}(p_r)| < |\vec{f}_{yr\to ys}(p_r)|$}
		\end{array}\right.
	\end{eqnarray}
	
	\begin{figure}[t]
		
		\begin{center}
			\includegraphics[width=0.94\linewidth]{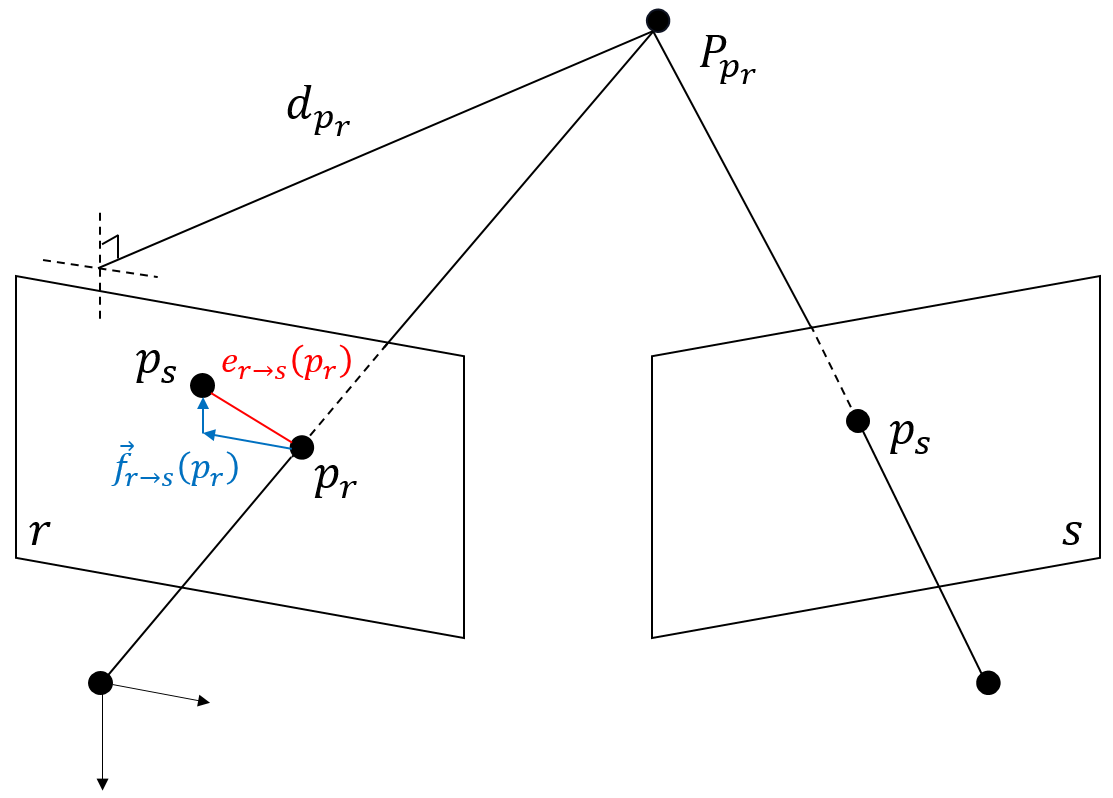}
		\end{center}
		
		\caption{
			\textbf{Depth ($d_{p_r}$), flow ($\vec{f}_{r \to s}(p_r)$) and E-flow ($e_{r \to s}(p_r)$).} We draw $p_s$ in $r$ to visualize the flow and E-flow. $d_{p_r}$ is the depth of $p_r$. $\vec{f}_{r \to s}(p_r)$ and $e_{r \to s}(p_r)$ describe the way $p_r$ moves. In static scenes, $d_{p_r}$, $\vec{f}_{r \to s}(p_r)$ and $e_{r \to s}(p_r)$ are all correlated and can be converted to each other based on multi-view geometry.
		}
		
		\label{fig:flow_depth_elow} 
		
	\end{figure}
	
	\subsection{E-flow: the epipolar disparity flow }
	
	The flow describes the movement of a pixel on the image plane but does not obey the epipolar geometry, which introduces ambiguity when triangulating the depth. Therefore, we use the epipolar geometry to restrain the flow and define the epipolar disparity flow (E-flow) by Eq. \ref{eq:elow}, where $\vec{e}_{dir}$ is the normalized direction vector of the epipolar line, and $ \cdot  $ is the dot product of vectors.
	
	\begin{eqnarray}
		\label{eq:elow}
		e_{r \to s}(p_r) = \vec{e}_{dir}(p_r) \cdot  ( p_s - p_r )
	\end{eqnarray}
	
	\textbf{E-flow and Flow} Compared with the flow, the E-flow is a scalar and only moves on the epipolar line. In the static scene, the flow and the E-flow are two different ways to describe pixel movement, and their relationship is shown in Eq. \ref{eq:elow_and_flow}. Fig. \ref{fig:flow_depth_elow} visualizes the flow $\vec{f}_{r \to s}(p_r)$ and the E-flow $e_{r \to s}(p_r)$ of pixel $p_r$.
	
	\begin{eqnarray}
		\label{eq:elow_and_flow}
		\vec{f}_{r \to s}(p_r) = \vec{e}_{dir}(p_r) e_{r \to s}(pr)
	\end{eqnarray}
	
	\textbf{E-flow and Depth} Considering the relationship between the E-flow and the flow, and the relationship between the flow and the depth, we can convert E-flow to depth, and vice versa by Eq. \ref{eq:elow_and_depth}, in which $\Leftrightarrow$ denotes the interconversion.
	
	\begin{eqnarray}
		\label{eq:elow_and_depth}
		e_{r \to s} \Leftrightarrow \vec{f}_{r \to s} \Leftrightarrow d_r
	\end{eqnarray}
	
	\begin{figure*}[t]
		
		\begin{center}
			\includegraphics[width=0.92\linewidth]{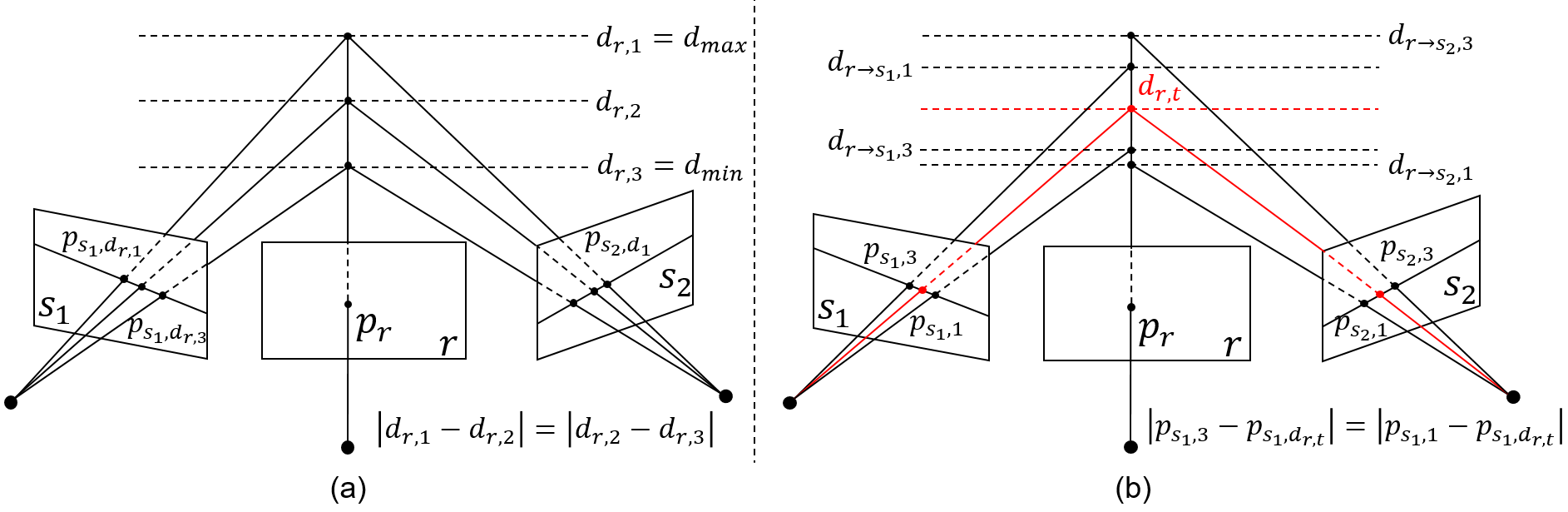}
		\end{center}
		
		\caption{
			\textbf{Different Sampling Spaces.} (a) shows the sampling space of existing methods that evenly samples in the 3D space to build a 3D cost volume. (b) shows the sampling space of DispMVS that evenly samples on each source image plane and estimates the depth by triangulation, whose distribution is decided by the relative poses of the source image.
		}
		
		\label{fig:sampling_space}
		\vspace{-1em}
		
	\end{figure*}
	
	\subsection{DispMVS}
	
	Given a reference image $r$ and $N$ source images $s_i(1<=i<=N)$, DispMVS firstly extracts features from all input images and then iteratively updates the depth from random initialization. DispMVS separately estimates the E-flow of each pair by a 2D cost volume and converts the E-flow to the depth for later multi-view depth fusion by a weighted sum. Fig. \ref{fig:pipeline} shows the pipeline of DispMVS with the first iteration at the coarse stage.
	
	\textbf{Feature Extraction} Following RAFT \cite{teed2020raft}, we use two identical encoders to extract features from input images. One encoder simultaneously extracts matching features from input images to calculate similarities. Meanwhile, another encoder extracts context features from the reference image to iteratively update the E-flow. As we apply a coarse-to-fine strategy to speed up efficiency and accuracy, these encoders use a UNet structure \cite{ronneberger2015u} to extract coarse feature maps $c_r,c_{s_i}$ with resolution $1/16$ for the coarse stage and $f_r,f_{s_i}$ with resolution $1/4$ for the fine stage. Besides, we also use the deformable convolutional network \cite{dai2017deformable} at the decoder part to capture valuable information.
	
	\textbf{Initialization of E-flow} DispMVS relies on the E-flow to estimate the depth, so it needs an initial depth as the starting point. We adopt the initialization strategy from PatchMatch \cite{bleyer2011patchmatch} and initialize $d_{r,0}$ by Eq. \ref{eq:random_init}, where $rand$ comes from a normal distribution, and $(d_{min},d_{max})$ are lower and upper bound of the depth range. With $d_{r,0}$, DispMVS can initialize the E-flow for pair by Eq. \ref{eq:elow_and_depth}, which is zero for the first iteration. Fig. \ref{fig:pipeline} shows how DispMVS reconstructs a coarse depth from a random depth.
	
	\begin{eqnarray}
		\label{eq:random_init}
		\frac{1}{d_{r,0}} = {rand\times(\frac{1}{d_{min}} - \frac{1}{d_{max}}) + \frac{1}{d_{max}}}
	\end{eqnarray}
	
	\textbf{E-flow Estimation} After the initialization, DispMVS estimates  $e_{s_i}$ of each pair $(r,s_i)$ by the GRU module. For $p_r$ and coarse feature $c_r(p_r)$, DispMVS uses $e_{r \to s_i}(p_r)$ to find matched point $p_{s_i}$ on $s_i$ and samples feature from $c_{s_i}(p_s)$. To increase the reception field, DispMVS applies $m_s$ times of the average-pool on $c_{s_i}$ to generate features with different scales and evenly samples $m_p$ points around $p_{s_i,t}$ along the epipolar line at each scale with a distance of one pixel. Then, DispMVS calculates the similarity between $c_r(p_r)$ and $m_p \times m_s $ features from $c_{s_i}$ by Eq. \ref{eq:simi} to build a 2D cost volume. In the end, DispMVS feeds the cost volume and $e_{r \to s_i}(p_r)$ to the GRU module to estimate a new $e_{r \to s_i}(p_r)$ and the weight $w_{s_i}(p_r)$. In DispMVS, we set $m_s=4,m_p=9$ at the coarse stage and $m_s=2,m_p=5$ at the fine stage. Fig. \ref{fig:sampling_space} compares two different sampling spaces. 
	
	\begin{eqnarray}
		\label{eq:simi}
		simi(c_r(p_r),c_{s_i}(p_{s_i})) = \sum_{0<=i<=D}c_r(p_r)[i]c_{s_i}(p_{s_i})[i]
	\end{eqnarray}

	\textbf{Multi-View E-flow Estimation} Since the E-flow only applies to two views, DispMVS utilizes a weighted sum to extend the E-flow to the multi-view situation. DispMVS converts $e_{r \to s_i}$ to $d_{r \to s_i}$ by Eq. \ref{eq:elow_and_depth} and then fuses $d_r$ by Eq. \ref{eq:depth_fuse}, where $w_{s_i}$ is normalized by the softmax. To be noted, DispMVS iteratively reconstruct the depth, which means that there are several conversion between the depth and the E-flow, thus further improving multi-view consistency.
	
	\begin{eqnarray}
		\label{eq:depth_fuse}
		\left\{\begin{array}{@{}l@{\quad}l}
			w_{s,t+1} &= softmax(w_{s,t+1}) \\[\jot]
			d_{r,t+1} &= \sum_{1<=i<=N} { d_{r \to s_i,t+1} \times w_{s_i,t+1} }  
		\end{array}\right.
	\end{eqnarray}
	
	\textbf{Coarse-to-Fine Stragety} Following CasMVSNet \cite{gu2020cascade}, DispMVS uses a coarse-to-fine strategy. DispMVS uses $c_r,c_{s_i}$ with $t_c$ iterations at the coarse stage and  $f_r,f_{s_i}$ with $t_f$ iterations at the fine stage. DispMVS starts from a random depth at the coarse stage and upsamples \cite{teed2020raft} the coarse depth to the fine stage for later refinement. Generally, DispMVS needs more $t_c$ and fewer 
	$t_f$ to improve efficiency and accuracy. 
	
	\textbf{Loss Function} As DispMVS outputs a depth in each iteration, we calculate the L1-loss for all depth maps to speed up convergence. To improve the stability of the training procedure, we use the inverse depth range to normalize the ground truth (gt) and the depth $d_{r,i}$. Eq. \ref{eq:loss} shows our loss function, in which $\gamma=0.9$:
	
	\begin{eqnarray}
		\label{eq:loss}
		loss = \sum_{j={t_c,t_f}}{\sum_{0<=i<j}{\gamma^{i}|norm(gt_i)-norm(d_{r,i})|}} 
	\end{eqnarray}

	\section{Experiments}
	
	In this section, we benchmark our DispMVS on two public datasets and compare it with a set of existing methods. We also conduct ablation experiments to explore the effects of different settings of DispMVS.
	\begin{table}[ht]
		\begin{center}
			\caption{\textbf{The evaluation results on DTUMVS \cite{aanaes2016large}}. The lower the Accuracy (Acc), Completeness (Comp), Overall and Mem (GPU Memory), the better. We split methods into three categories and highlight the best in bold for each.}
			\label{tab:dtu_results}
			\addtolength{\tabcolsep}{-0.2pt}
			\resizebox{\linewidth}{!}{\begin{tabular}{c|lcccc}
					\hline\noalign{\smallskip}
					& Method & Acc$\downarrow$ & Comp$\downarrow$ & Overall$\downarrow$ & Mem$\downarrow$ \\
					\noalign{\smallskip}
					\hline
					\noalign{\smallskip}
					
					\multirow{4}{*}{Trad}
					&Comp \cite{campbell2008using}          & 0.835 & \textbf{0.554} & 0.695 & ---  \\
					&Furu \cite{furukawa2009accurate}          & 0.613 & 0.941 & 0.777 & ---  \\
					&Tola \cite{tola2012efficient}          & 0.342 & 1.190 & 0.766 & ---  \\
					&Gipuma \cite{galliani2015massively}        & \textbf{0.283} & 0.873 & \textbf{0.578} & ---  \\

					\hline
					
					\multirow{12}{*}{Conv}
					&MVSNet \cite{yao2018mvsnet}        & 0.396 & 0.527 & 0.462 & 9384M   \\
					&Point-MVSNet \cite{chen2019point}  & 0.342 & 0.411 & 0.376 & ---   \\
					&P-MVSNet \cite{luo2019p}      & 0.406 & 0.434 & 0.420 & ---  \\
					&Fast-MVSNet \cite{yu2020fast}   & 0.336 & 0.403 & 0.370 & ---   \\
					&CVP-MVSNet \cite{yang2020cost}    & \textbf{0.296} & 0.406 & 0.351 & ---  \\
					&Vis-MVSNet \cite{zhang2020visibility}    & 0.369 & 0.361 & 0.365 & 4775M   \\
					&CIDER \cite{xu2020learning}         & 0.417 & 0.437 & 0.427 & ---  \\
					
					&CasMVSNet \cite{gu2020cascade}     & 0.325 & 0.385 & 0.355 & 4591M  \\
					&UCS-Net \cite{cheng2020deep} & 0.338 & 0.349 & 0.344 & 4057M \\
					&EPP-MVSNet \cite{ma2021epp}    & 0.414 & 0.297 & 0.355 & ---  \\
					&UniMVSNet \cite{peng2022rethinking} & 0.352 & 0.278 & 0.315 & 3216M \\
					&GBiNet \cite{mi2022generalized}    & 0.315 & \textbf{0.262} & \textbf{0.289} & \textbf{2018M}  \\
					
					\hline
					
					\multirow{7}{*}{RNN}
					&R-MVSNet \cite{yao2019recurrent}      & 0.383 & 0.452 & 0.417 & ---  \\
					&D2HC-RMVSNet \cite{yan2020dense}  & 0.395 & 0.378 & 0.386 & ---  \\
					&AA-RMVSNet \cite{wei2021aa}    & 0.376 & 0.339 & 0.357 & 11973M \\
					&PatchMatchNet \cite{wang2021patchmatchnet} & 0.427 & \textbf{0.277} & 0.352 & 1629M  \\
					&IterMVS \cite{wang2022itermvs}       & 0.373 & 0.354 & 0.363 & \textbf{842M}  \\
					&DispMVS (ours)     & \textbf{0.354} & 0.324 & \textbf{0.339} & 1368M  \\
					
					\hline
			\end{tabular}}
		\end{center}
		\vspace{-1em}
	\end{table}
	\begin{figure}[t]
		
		\begin{center}
			\includegraphics[width=1.0\linewidth]{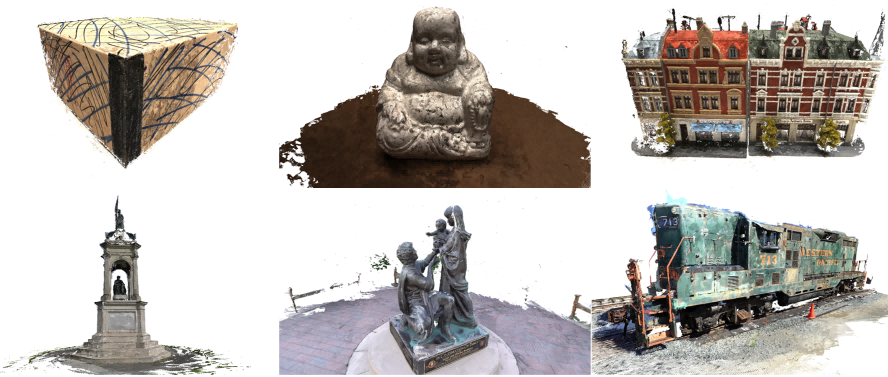}
		\end{center}
		
		\caption{
			\textbf{Point clouds}  We visualize some point clouds generated by DispMVS on DTUMVS \cite{aanaes2016large} and Tanks \& Temple \cite{knapitsch2017tanks}.
		}
		
		\label{fig:viz_results}
		\vspace{-1em}
	\end{figure}
	
	\begin{table*}
		\scriptsize
		\begin{center}
			\caption{ \textbf{The evaluation results on Tanks \& Temple \cite{knapitsch2017tanks}}. Higher scores indicate higher quality of the point cloud. We split methods into two categories and highlight the best in bold for each. }
			\label{tab:tanks_result}
			\resizebox{\linewidth}{!}{\begin{tabular}{lccccccc|ccccccccc}
					\hline\noalign{\smallskip}
					\multicolumn{8}{c}{Advanced} &  \multicolumn{9}{|c}{Intermediate} \\
					\hline
					\noalign{\smallskip}
					Method & Mean & Aud. & Bal. & Cou. & Mus. & Pal. & Tem. & Mean & Fam. & Fra. & Hor. & Lig. & M60 & Pan. & Pla. & Tra \\
					\noalign{\smallskip}
					
					\hline
					
					MVSNet \cite{yao2018mvsnet} & - & - & - & - & - & - & - & 43.48 & 55.99 & 28.55 & 25.07 & 50.79 & 53.96 & 50.86 & 47.90 & 34.69 \\
					
					Point-MVSNet \cite{chen2019point} & - & - & - & - & - & - & - & 48.27 & 61.79 & 41.15 & 34.20 & 50.79 & 51.97 & 50.85 & 52.38 & 43.06 \\
					
					UCS-Net \cite{cheng2020deep} & - & - & - & - & - & - & - & 54.83 & 76.09 & 53.16 & 43.03 & 54.00 & 55.60 & 51.49 & 57.38 & 47.89 \\
					
					Vis-MVSNet \cite{zhang2020visibility}  & 33.78 & 20.79 & 38.77 & 32.45 & 44.20 & 28.73 & 37.70 & 60.03 & 77.40 & 60.23 & 47.07 & 63.44 & 62.21 & 57.28 & 60.54 & 52.07 \\
					
					CasMVSNet \cite{gu2020cascade} & 31.12 & 19.81 & 38.46 & 29.10 & 43.87 & 27.36 & 28.11 & 56.42 & 76.36 &  58.45 & 46.20 & 55.53 & 56.11 & 54.02 & 58.17 & 46.56 \\
					
					EPP-MVSNet \cite{ma2021epp} & 35.72 & 21.28 & 39.74 & 35.34 & 49.21 & 30.00 & 38.75 & 61.68 & 77.86 & 60.54 & 52.96 & 62.33 & 61.69 & 60.34 & 62.44 & 55.30 \\
					
					UniMVSNet \cite{peng2022rethinking} & \textbf{38.96} & 28.33 & \textbf{44.36} & \textbf{39.74} & \textbf{52.89} & \textbf{33.80} & 34.63 & \textbf{64.36} & \textbf{81.20} & 66.43 & \textbf{53.11} & \textbf{63.46} & \textbf{66.09} & \textbf{64.84} & \textbf{62.23} & \textbf{57.53} \\
					
					GBiNet \cite{mi2022generalized} & 37.32 & \textbf{29.77} & 42.12 & 36.30 & 47.69 & 31.11 & \textbf{36.93} & 61.42 & 79.77 & \textbf{67.69} & 51.81 & 61.25 & 60.37 & 55.87 & 60.67 & 53.89 \\
					
					\hline
					
					R-MVSNet \cite{yao2019recurrent} & 24.91 & 12.55 & 29.09 & 25.06 & 38.68 & 19.14 & 24.96 & 48.40 & 69.96 & 46.65 & 32.59 & 42.95 & 51.88 & 48.80 & 52.00 & 42.38 \\
					
					D2HC-RMVSNet \cite{yan2020dense} & - & - & - & - & - & - & - & 59.20 & 74.69 & 56.04 & 49.42 & 60.08 & 59.81 & 59.61 & 60.04 & 53.92 \\
					
					AA-RMVSNet \cite{wei2021aa} & 33.53 & 20.96 & \textbf{40.15} & 32.05 & \textbf{46.01} & \textbf{29.28} & 32.71 & \textbf{61.51} & \textbf{77.77} & 59.53 & 51.53 & \textbf{64.02} & \textbf{64.05} & \textbf{59.47} & \textbf{60.85} & \textbf{54.90} \\
					
					PatchmatchNet \cite{wang2021patchmatchnet} & 32.31 & 23.69 & 37.73 & 30.04 & 41.80 & 28.31 & 32.29 & 53.15 & 66.99 & 52.64 & 43.24 & 54.87 & 52.87 & 49.54 & 54.21 & 50.81 \\
					
					IterMVS \cite{wang2022itermvs} & 33.24 & 22.95	& 38.74	& 30.64	& 43.44	& 28.39	& 35.27 & 56.94 & 76.12	& 55.80	& 50.53	& 56.05	& 57.68	& 52.62	& 55.70	& 50.99 \\
					
					DispMVS (ours) & \textbf{34.90} & \textbf{26.09} & 38.01 & \textbf{33.19} & 44.90 & 28.49 & \textbf{38.75} & 59.07 & 74.73 & \textbf{60.67} & \textbf{54.13} & 59.58 & 58.02 & 53.39 & 58.63 & 53.42 \\ 
					
					\hline
					
			\end{tabular}}
		\end{center}
		\vspace{-2em}
	\end{table*}
	
	\paragraph{Datasets}
	
	We use three different datasets throughout our experiments. The DTUMVS \cite{aanaes2016large} is an indoor dataset in a controlled environment containing 79 scenes for training, 22 for testing, and 18 for validation. The BlendedMVS \cite{yao2020blendedmvs} is a large dataset captured from various outdoor scenes, with 106 scenes for training and the rest 7 scenes for testing. The Tanks\&Temple \cite{knapitsch2017tanks} is an outdoor multi-view stereo benchmark that contains 14 real-world scenes under complex conditions.
	
	\paragraph{Implementation}
	
	We implement DispMVS by PyTorch \cite{paszke2019pytorch} and train two models on the DTUMVS and the BlendedMVS separately. On the DTUMVS, we set the image resolution to $640 \times 512$ and $N=5$. On the BlendedMVS, we set the image resolution to $768 \times 576$ and $N=5$. For all models, we apply the training strategy in PatchmatchNet \cite{wang2021patchmatchnet} for better learning of the weight and use the Adam \cite{kingma2015adam}( $\beta_1=0.9,\beta_2=0.999$ ) optimizer with an initial learning rate of 0.0002 that halves every four epochs for 16 epochs. The training procedure is finished on two V100 with $t_c=8, t_f=2$ considering the GPU memory limitation. During the evaluation, we filter and fuse all depth maps into point clouds to compare with the ground truth.
	
	\subsection{Evaluation on DTUMVS} 
	We evaluate the DTUMVS on the test part and resize all images to $1600 \times 1152$ with $N=5$. Table \ref{tab:dtu_results} compares DispMVS with other state-of-the-art methods. We split existing methods into traditional methods, 3D convolution-based methods, and RNN-based methods. DispMVS has the best overall score among RNN-based methods and is 0.024 lower than IterMVS \cite{wang2022itermvs} and 0.018 lower than AA-RMVSNet \cite{wei2021aa}. DispMVS ranks the 3rd among all methods. GBiNet \cite{mi2022generalized} and UniMVSNet \cite{peng2022rethinking} are the top two methods, but they incur much higher GPU memory. We visualize some point clouds of DTUMVS generated by DispMVS in the first row of Fig. \ref{fig:viz_results}. These qualitative and quantitative experimental results demonstrate the effectiveness of DispMVS in obtaining depth by triangulation, even though DispMVS does not construct any 3D cost volume.
	
	\subsection{Evaluation on Tanks\&Tmple } 
	As Tanks\&Temple does not provide training samples, we apply the model pretrained on the BlendedMVS to it. We resize all images of Tanks\&Temple to $1600 \times 1152 $ and set $N=7$. Table \ref{tab:tanks_result} shows the results of learning-based methods, split into 3D convolution-based and RNN-based. Tanks\&Temple contains two subsets, Intermedia and Advanced. DispMVS achieves the best mean score on the Advanced subset among the RNN-based method. Although AA-RMVSNet \cite{wei2021aa} outperforms DispMVS on the Intermedia subset, AA-RMVSNet uses nearly ten times more GPU memory than DispMVS. Overall, the results on Tanks\&Temple demonstrate that DispMVS has robust generalization with a low amount of GPU memory. The second row of Fig. \ref{fig:viz_results} shows point clouds generated on the Tanks\&Temple by DispMVS.
	
	\subsection{Ablation Studies}
	
	In this subsection, we discuss the core parts of our method. Considering that the RAFT structure has been thoroughly studied in \cite{teed2020raft,lipson2021raft,wang2022itermvs}, we conduct ablation experiments on the coarse-to-fine strategy, the random initialization, and the changes in depth range. Throughout all ablation experiments, we use the DTUMVS as a baseline dataset. 
	
	\textbf{The coarse-to-fine stragety} DispMVS-M firstly estimates the depth with the feature from $1/16$ and refines the depth with the feature from $1/4$. DispMVS-S only extracts features from $1/8$ to recover the depth map to make the comparison fairer. Table. \ref{tab:dtu_stage} shows that the coarse-to-fine strategy dramatically improves the overall score from 0.455 for DispMVS-M to 0.339 for DispMVS-S. Therefore, we choose DispMVS-M as the method used in this paper.

	\textbf{Random Initilization} Unlike existing methods that build the 3D cost volume from a known depth range, DispMVS starts from a random initial depth, which means that the input of DispMVS could be different every time. To measure the effects of the random initialization, we conduct three times of initilization without fixing the random seed and evaluate point clouds. Table \ref{tab:dtu_random} shows that the variance of metrics between different inferences is smaller than 1e-6, which proves that DispMVS are robust to the random initial depth and can always generate a high-quality depth map. 
	
	\begin{table}[ht]
		\begin{center}
			\caption{\textbf{Comparison between single stage and multi stage}. 
				``-S" indicates reconstruction with only one resolution at $1/8$, while ``-M" indicates reconstruction with multiple resolutions (the method in this paper).}
			\label{tab:dtu_stage}
			\addtolength{\tabcolsep}{-0.2pt}
			\begin{tabular}{l|ccc}
				\hline\noalign{\smallskip}
				Method & Acc$\downarrow$ & Comp$\downarrow$ & Overall$\downarrow$ \\
				\noalign{\smallskip}
				\hline
				\noalign{\smallskip}
				
				DispMVS-S    & 0.500 & 0.410 & 0.455    \\
				DispMVS-M    & \textbf{0.354} & \textbf{0.324} & \textbf{0.339}    \\
				
				\hline
			\end{tabular}
		\end{center}
		\vspace{-2.2em}
	\end{table}
	
	\textbf{Depth Range} The existing MVS methods split a given depth range into several bins and build a 3D cost volume by differentiable homography. However, DispMVS is insensitive to the depth range as it only constructs a 2D cost volume on the image plane along the epipolar line. We select two state-of-the-art methods (GBiNet \cite{mi2022generalized} and IterMVS \cite{wang2022itermvs}) and manually change the depth range by Eq. \ref{eq:change_depth_range}. All methods are trained by the same dataset with the same depth range.
	Table. \ref{tab:dtu_range} shows that performance of GBiNet and IterMVS decreases dramatically, but DispMVS can be robust to these changes. Fig. \ref{fig:dtu_range} visualizes point clouds generated by different methods with different depth ranges, where GBiNet and IterMVS cannot converge when the depth range is too large.
	
	\begin{eqnarray}
		\label{eq:change_depth_range}
		range_x = \left\{\begin{array}{@{}l@{\quad}l}
			d_{min} &= d_{min}/x \\[\jot]
			d_{max} &= d_{max} \times x 
		\end{array}\right.
		\vspace{-1em}
	\end{eqnarray}
	
	\subsection{Limitations}
	
	As DispMVS needs to keep building the 2D cost volume during the iteration, its computational efficiency is relatively low. In our experiment, DispMVS needs around 0.7 seconds to process a view on the DTUMVS. Compared with IterMVS \cite{wang2022itermvs} which only needs around 0.3 seconds per view, DispMVS needs a more efficient epipolar matching module.
	In addition, DispMVS needs around 48G GPU memory during training because DispMVS needs several iterations to update the depth by the GRU module, which needs to save all gradients and intermediate results.
	
	\begin{table}[ht]
		\begin{center}
			\caption{\textbf{Evaluate the random initialization.} A lower variance means the difference between multi results is smaller.}
			\label{tab:dtu_random}
			\addtolength{\tabcolsep}{-0.2pt}
			\begin{tabular}{c|ccc}
				\hline\noalign{\smallskip}
				Method & Acc$\downarrow$ & Comp$\downarrow$ & Overall$\downarrow$ \\
				\noalign{\smallskip}
				\hline
				\noalign{\smallskip}
				
				rand-1     & 0.353829 & 0.324110 & 0.338970    \\
				rand-2     & 0.354811 & 0.324946 & 0.339878    \\
				rand-3     & 0.354272 & 0.324324 & 0.339298    \\
				
				\hline
				
				variance   & \textbf{2.418e-7} & \textbf{1.890e-7} & \textbf{2.110e-7}    \\
				\hline
			\end{tabular}
		\end{center}
		\vspace{-1.5em}
	\end{table}
	\begin{table}[ht]
		\begin{center}
			\caption{ \textbf{Influences of changing the depth range.} The lower, the better for all metrics under different depth ranges.  }
			\label{tab:dtu_range}
			\addtolength{\tabcolsep}{-0.2pt}
			\resizebox{\linewidth}{!}{\begin{tabular}{c|l|ccc}
					\hline\noalign{\smallskip}
					Range & Method & Acc$\downarrow$ & Comp$\downarrow$ & Overall$\downarrow$ \\
					\noalign{\smallskip}
					\hline
					\noalign{\smallskip}
					
					\multirow{3}{*}{$range_1$}
					&GBiNet \cite{mi2022generalized}     & \textbf{0.315} & \textbf{0.262} & \textbf{0.289}    \\
					&IterMVS \cite{wang2022itermvs}    & 0.373 & 0.354 & 0.363    \\
					&DispMVS (ours)        & 0.354 & 0.324 & 0.339    \\
					
					\hline
					
					\multirow{3}{*}{$range_2$}
					&GBiNet \cite{mi2022generalized}     & 0.480 & 0.556 & 0.518    \\
					&IteMVS \cite{wang2022itermvs}    & 0.532 & 1.471 & 1.002    \\
					&DispMVS (ours)        & \textbf{0.348} & \textbf{0.404} & \textbf{0.376}    \\
					
					\hline
					
					\multirow{3}{*}{$range_3$}
					&GBiNet \cite{mi2022generalized}      & 0.618 & 1.303 & 0.960    \\
					&IterMVS \cite{wang2022itermvs}    & 0.935 & 6.985 & 3.960    \\
					&DispMVS (ours)         & \textbf{0.314} & \textbf{0.671} & \textbf{0.493}    \\
					
					\hline
			\end{tabular}}
		\end{center}
		\vspace{-1.8em}
	\end{table}

	\section{Conclusion}
	
	This paper introduces a new pipeline of MVS, called DispMVS, which does not need to build any 3D cost volumes but triangulates the depth map by multi-view geometry. DispMVS is a depth range invariant method and can be generalized to the dataset with different ranges with the training set, which proves that 3D cost volume is unnecessary for MVS. Compared with existing learning-based methods, DispMVS lets the network focus on matching pixels that the CNN network is good at and uses the multi-view geometry to deal with the geometry information. Experiments on datasets show that DispMVS achieves comparable results with other 3D convolution methods and outperforms RNN-based methods with a lower GPU memory requirement.
	
	\paragraph{Acknowledgments}
	This work was supported in part by grant RMGS2021-8-10 from Hong Kong Research Matching Grant Scheme and the NVIDIA Academic Hardware Grant.

	\newpage
	\clearpage
	
	\begin{center}
		{\Huge \bf Supplementary material}
	\end{center}
	
	\setcounter{section}{0}
	
	\section{Depth Normalization}
	
	To make our method more numeric stable, we apply a depth normalization by Eq. \ref{eq:normalize}, where we use the inversed minimum depth $d_{min}$ and inversed maximum depth $d_{max}$ to normalize the input depth. The depth normalization can decrease the effect of outliers, especially in the multi-view depth fusion and the loss function. In the multi-view depth fusion, we convert the epipolar disparity flow (E-flow) to depth by triangulation, which may introduce outliers when E-flow is too small. As for the loss function, the normalization can make it easier for the network to converge when the depth range is different among datasets.

	\begin{eqnarray} 
		\label{eq:normalize}
		depth\_norm = \frac{1/depth - 1/depth_{max}}{1/depth_{min} - 1/depth_{max}}
	\end{eqnarray}
	
	\section{GRU Update}
	
\begin{figure}
		
		\includegraphics[width=1.0\linewidth]{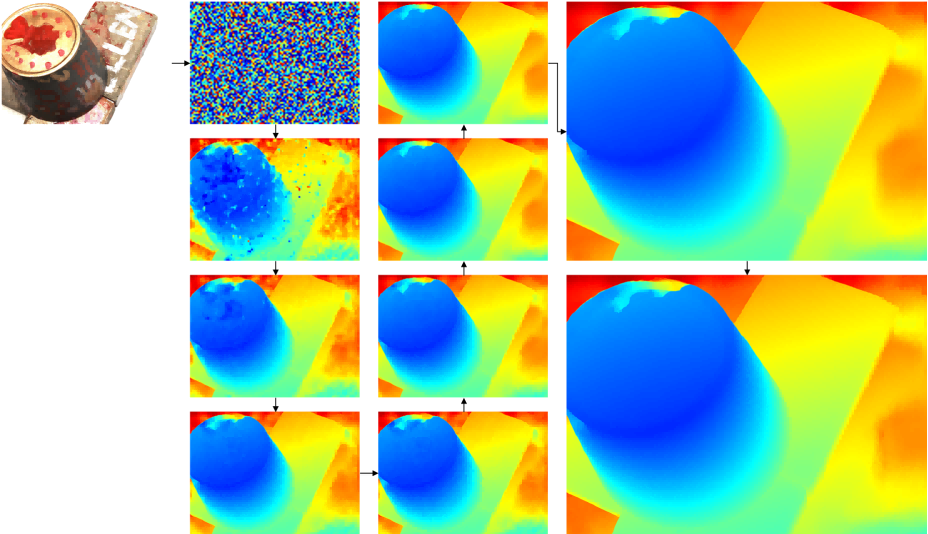}

		\includegraphics[width=1.0\linewidth]{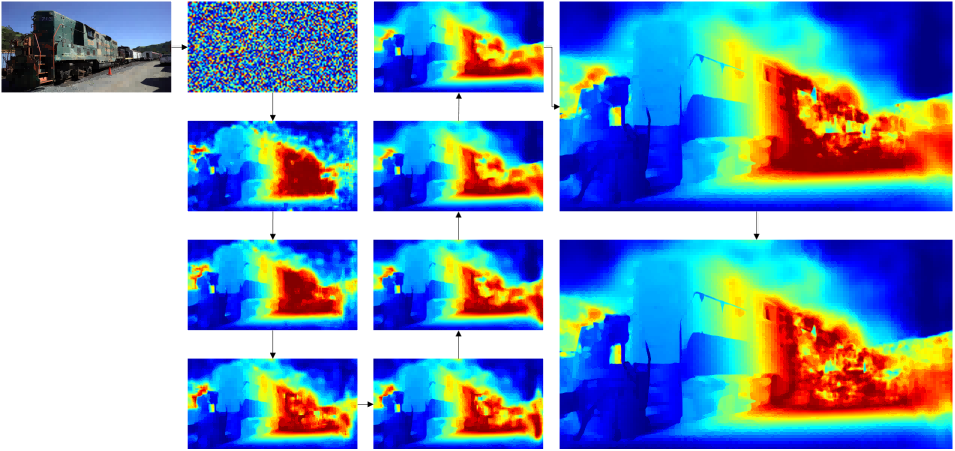}

		\caption{
			\textbf{Depth Maps in each iteration on DTUMVS and Tanks\&Temple} DispMVS iteratively recovers the depth map. An interesting phenomenon is that DispMVS cannot reconstruct the sky region, where DispMVS cannot find reliable matches and is far away from the view. But DispMVS can reconstruct the thin structure nearby.
		}
		
		\label{fig:dtu_iter}
		
	\end{figure}

	To update the E-flow, we utilize two GRU modules at the coarse and fine stages of DispMVS with the same network structure of RAFT \cite{teed2020raft}. The input and output of this module is seen in Eq. \ref{eq:gru_update}, where $f_t,f_r,f_h$ are 2D convolution networks and $W_t,W_r,W_h$ are the corresponding parameters. After updating E-flow, we triangulate the depth and fuse them in each iteration. Fig. \ref{fig:dtu_iter} shows the depth maps generated by DispMVS on DTU and Tanks\&Temple, in which we can see that the depth map recovers from coarse to fine.
	
	\begin{eqnarray}
		\label{eq:gru_update}
		\left\{\begin{array}{@{}l@{\quad}l}
			z_t &= \sigma( f_t([h_{t-1},x_t],W_t) ) \\[\jot]
			r_t &= \sigma( f_r([h_{t-1},x_t],W_r) ) \\[\jot]
			\tilde{h_t} &= tanh( f_h([r_t \bigodot h_{t-1},x_t],W_h) ) \\[\jot]
			h_t &= (1-z_t)\bigodot h_{t-1} + z_t \bigodot \tilde{h_t}
		\end{array}\right.
	\end{eqnarray}
	
	\section{Depth Upsampling}
	
	Following RAFT \cite{teed2020raft}, we upsample a coarse depth map to a fine depth map through a convex combination of each pixel's nine neighbors instead of directly using the bilinear upsampling. Since this upsampling module acquires the fusion weights by learning, it can retain more details and have sharp edges at the boundaries. 
	
	\section{Depth Fusion}
	
\begin{figure*}
		\centering

		\includegraphics[width=.24\linewidth]{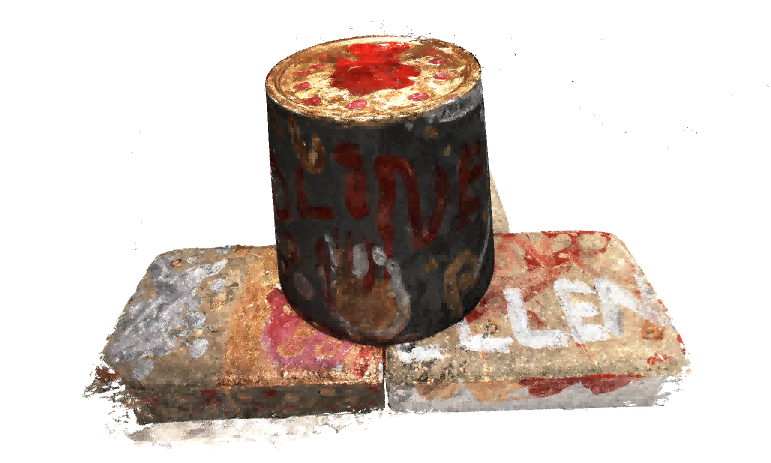}
		\includegraphics[width=.24\linewidth]{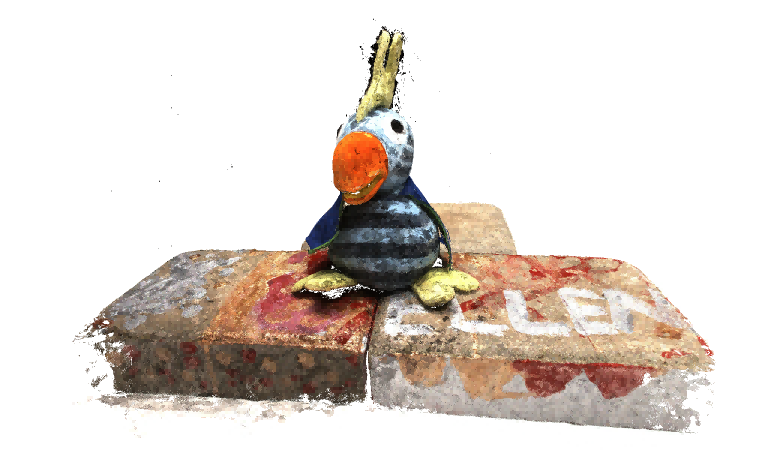}
		\includegraphics[width=.24\linewidth]{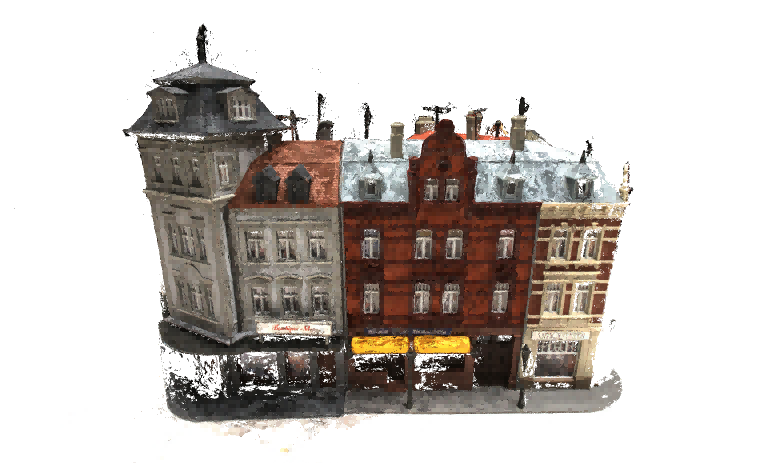}
		\includegraphics[width=.24\linewidth]{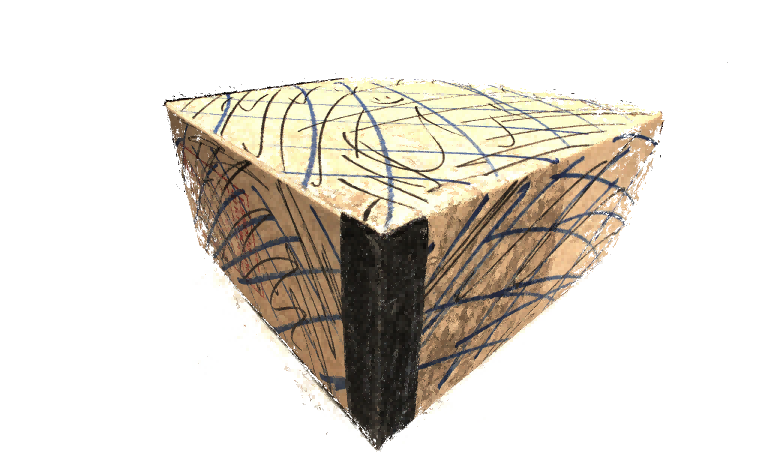}
		
		\includegraphics[width=.24\linewidth]{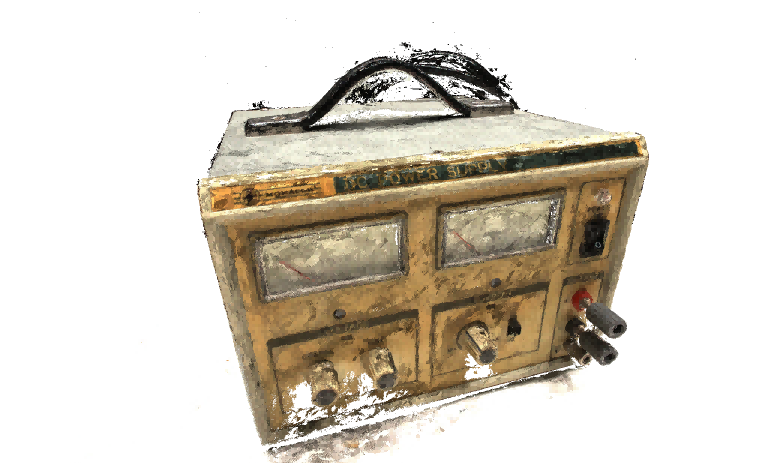}
		\includegraphics[width=.24\linewidth]{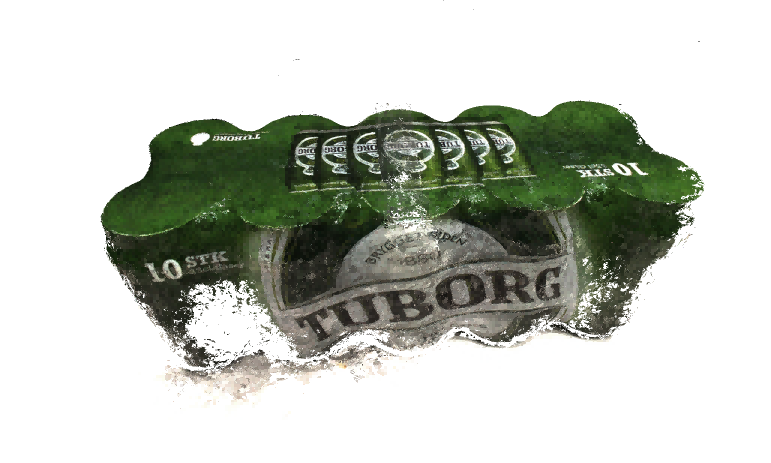}
		\includegraphics[width=.24\linewidth]{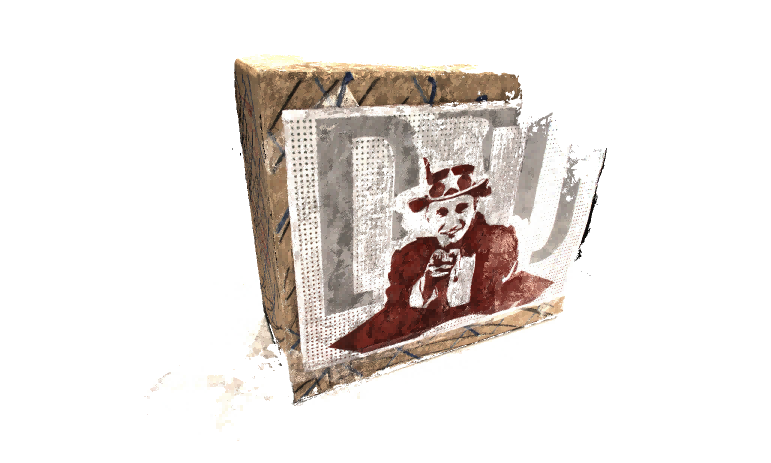}
		\includegraphics[width=.24\linewidth]{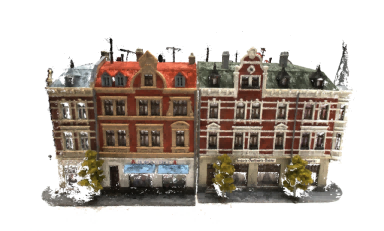}
		
		\includegraphics[width=.24\linewidth]{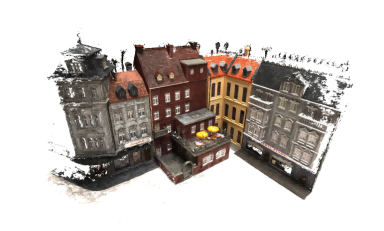}
		\includegraphics[width=.24\linewidth]{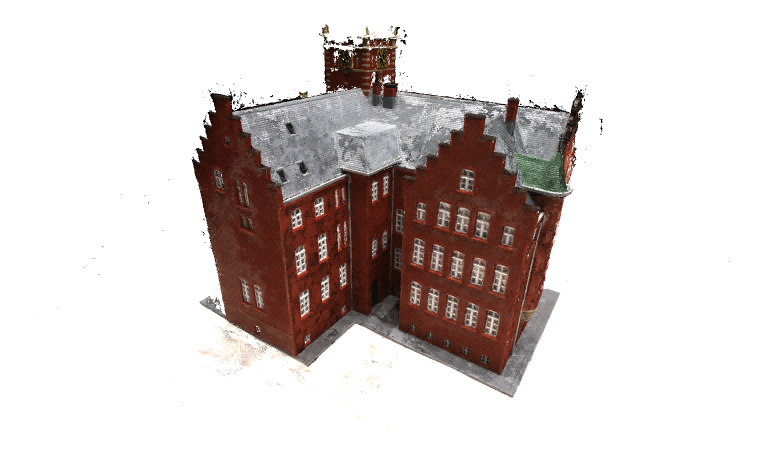}
		\includegraphics[width=.24\linewidth]{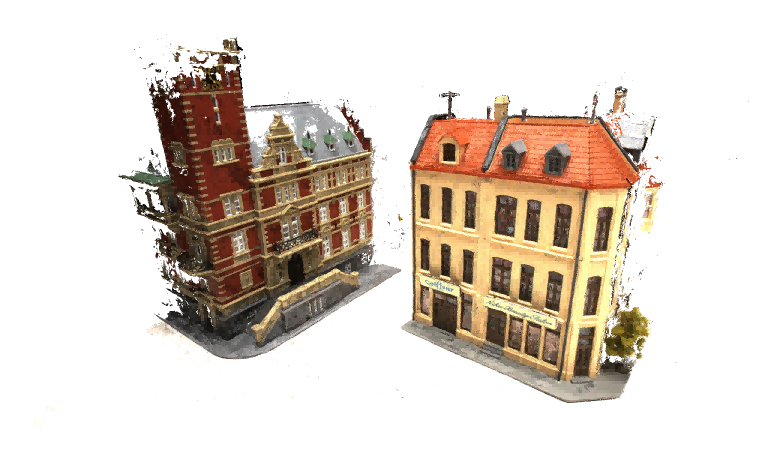}
		\includegraphics[width=.24\linewidth]{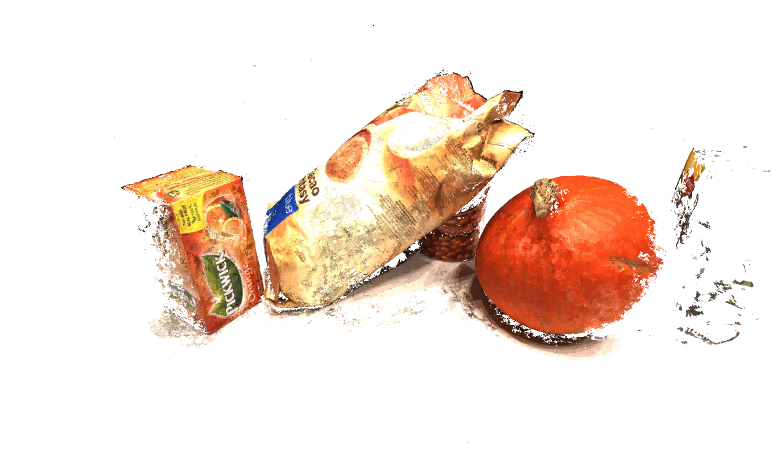}
		
		\includegraphics[width=.24\linewidth]{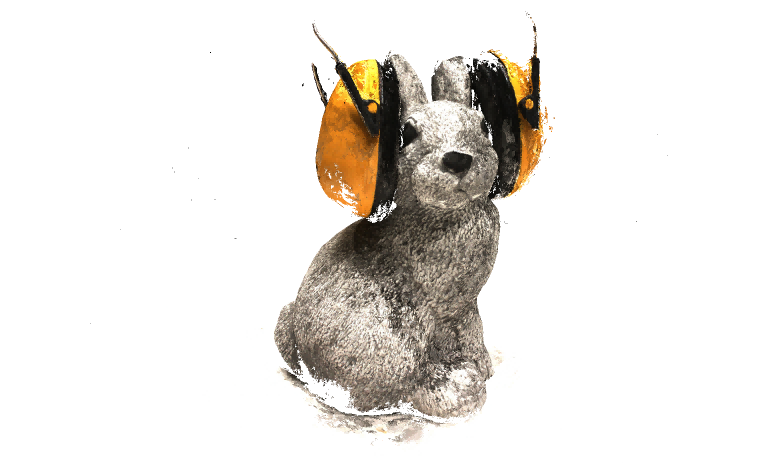}
		\includegraphics[width=.24\linewidth]{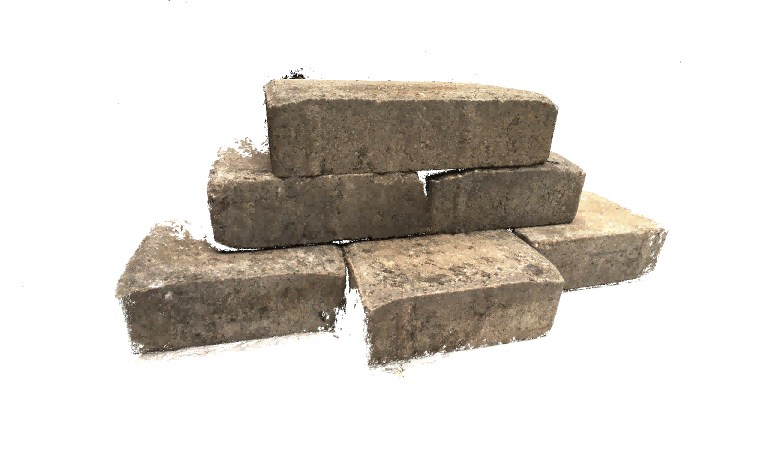}
		\includegraphics[width=.24\linewidth]{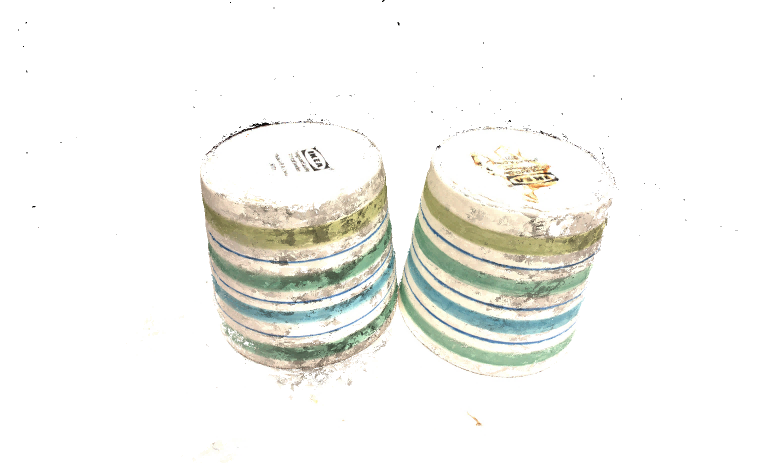}
		\includegraphics[width=.24\linewidth]{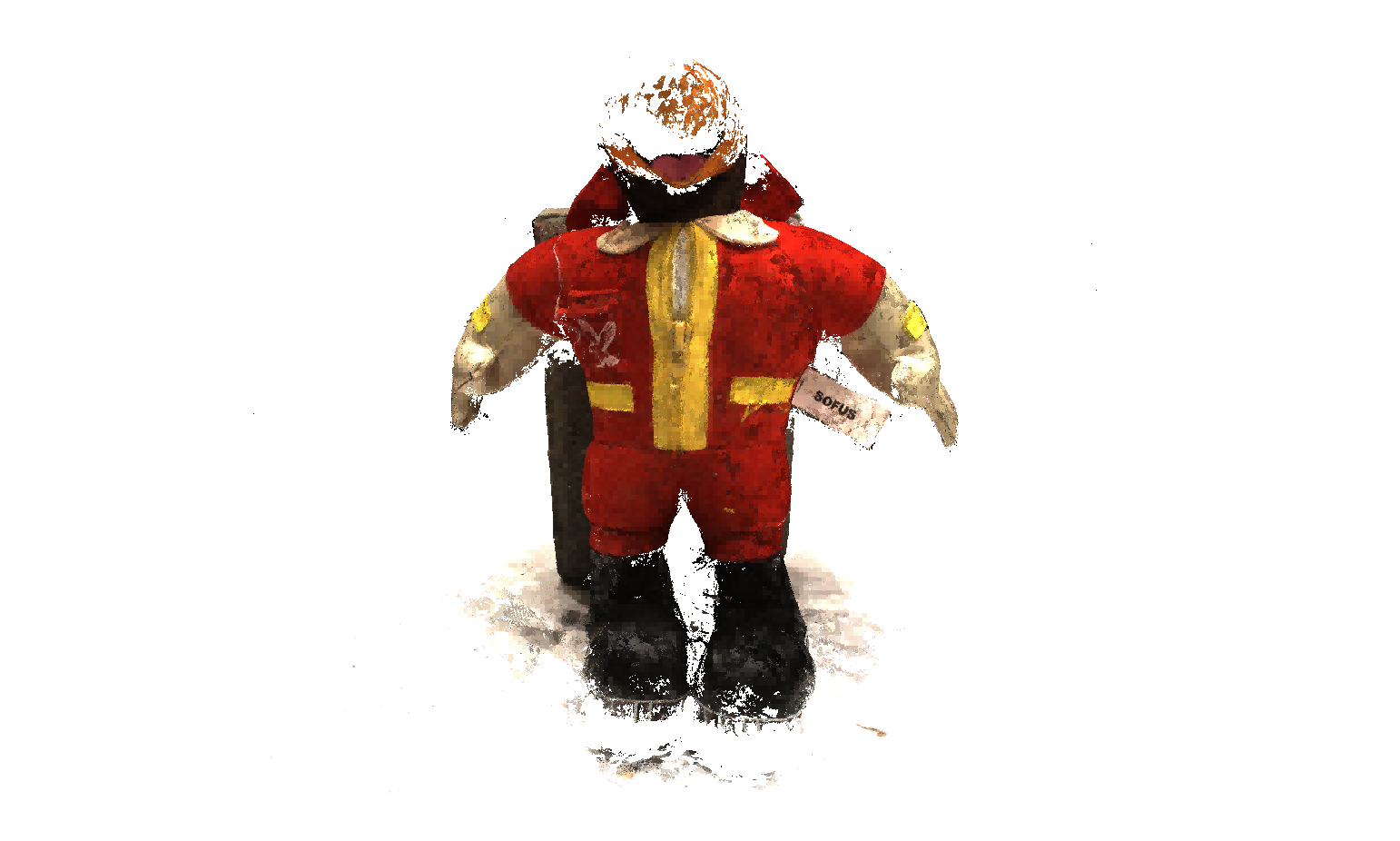}
		
		\includegraphics[width=.24\linewidth]{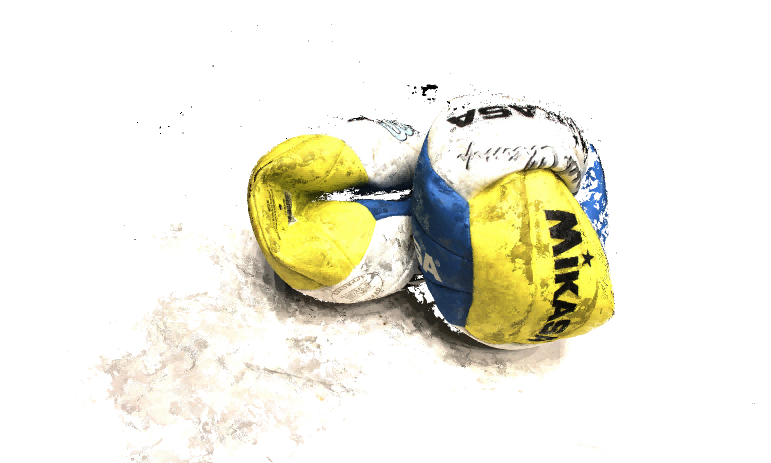}
		\includegraphics[width=.24\linewidth]{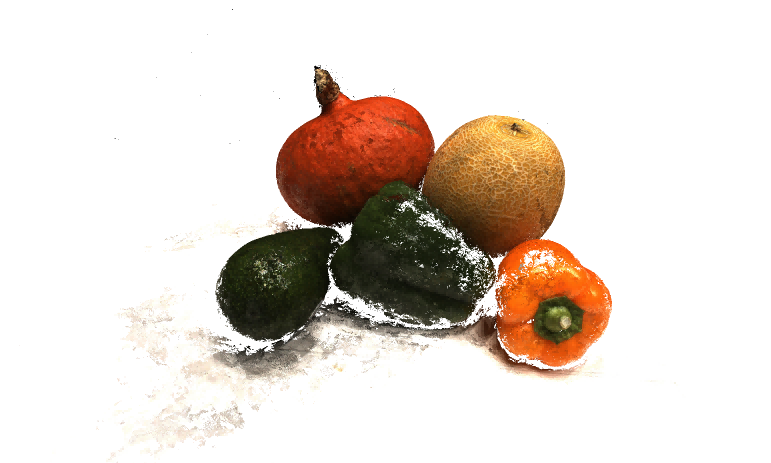}
		\includegraphics[width=.24\linewidth]{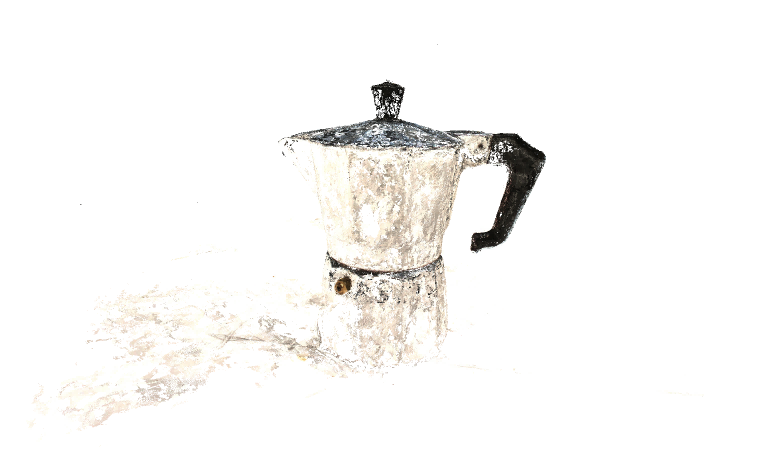}
		\includegraphics[width=.24\linewidth]{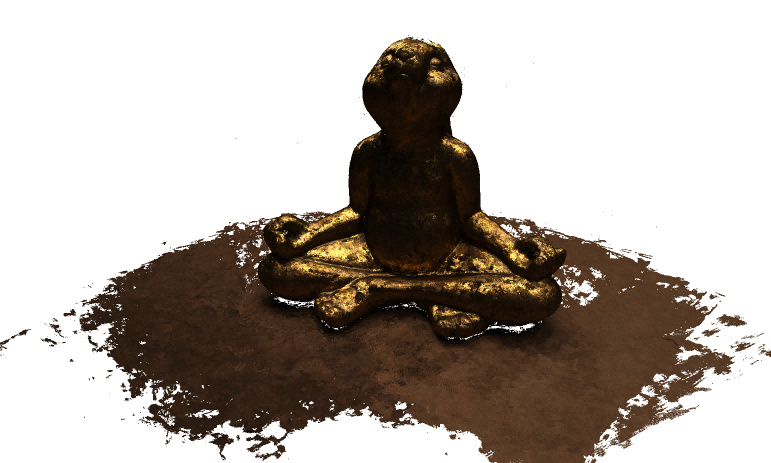}
		
		\includegraphics[width=.4\linewidth]{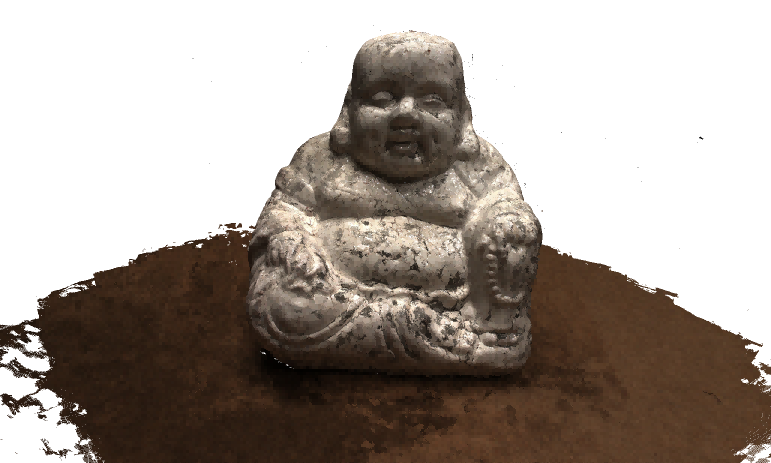}
		\includegraphics[width=.4\linewidth]{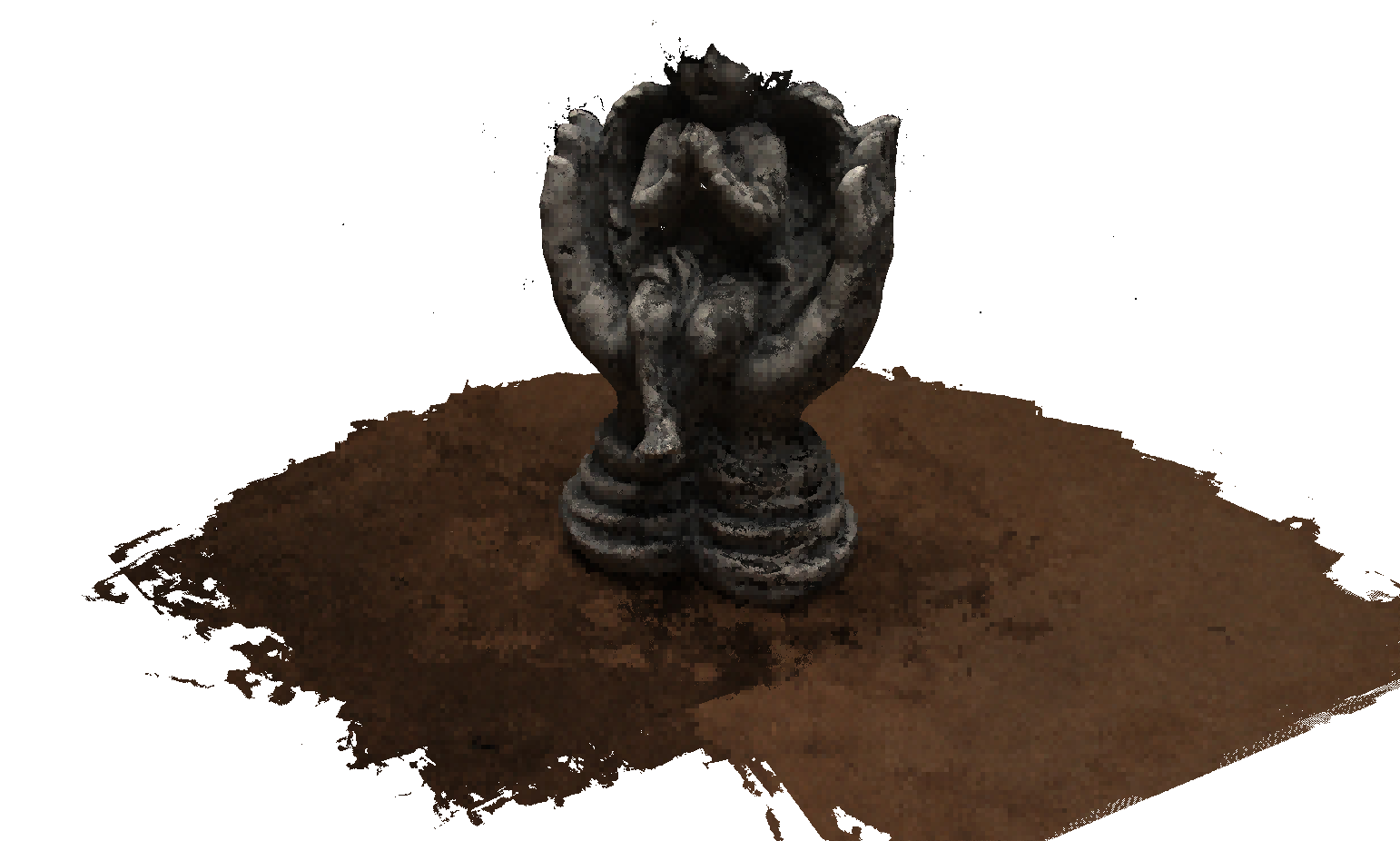}
		
		\caption{
			\textbf{Results on DTUMVS} We visualize 22 scenes in the test.
		}
		
		\label{fig:dtu_pointcloud}
	\end{figure*}

\begin{figure*}
		\centering

		\includegraphics[width=.3\linewidth]{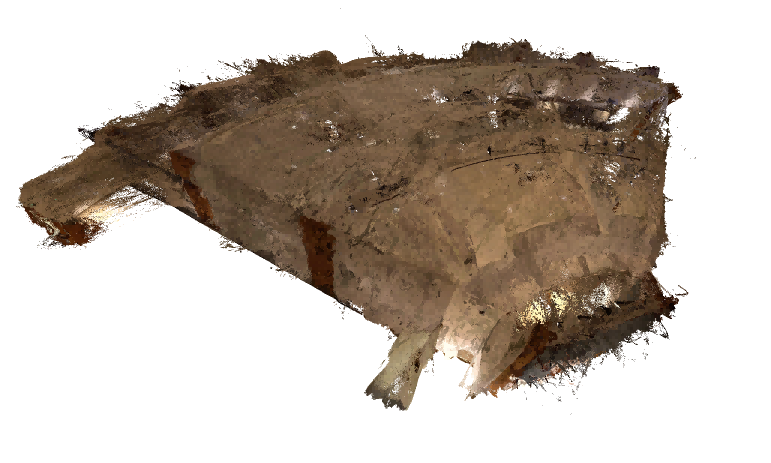}
		\includegraphics[width=.3\linewidth]{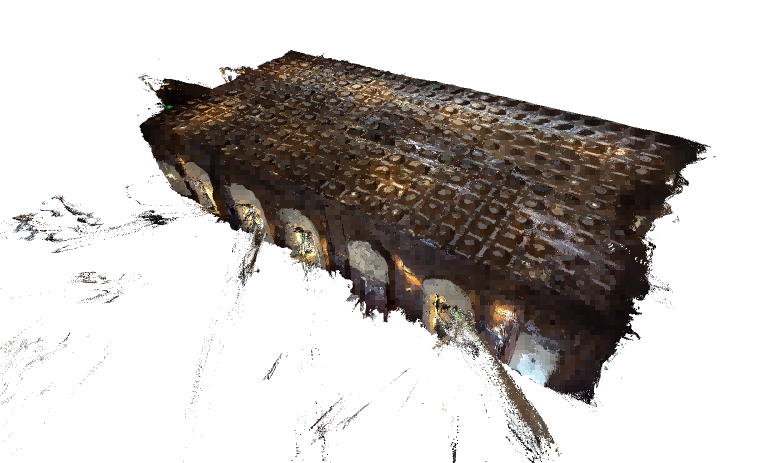}
		\includegraphics[width=.3\linewidth]{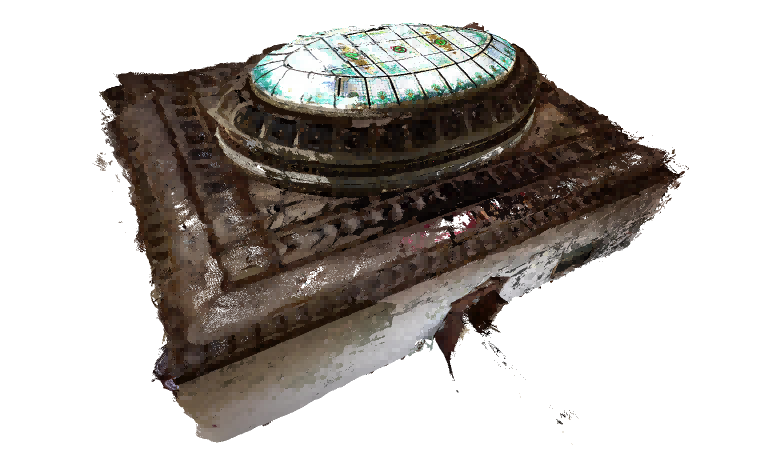}
		
		\includegraphics[width=.3\linewidth]{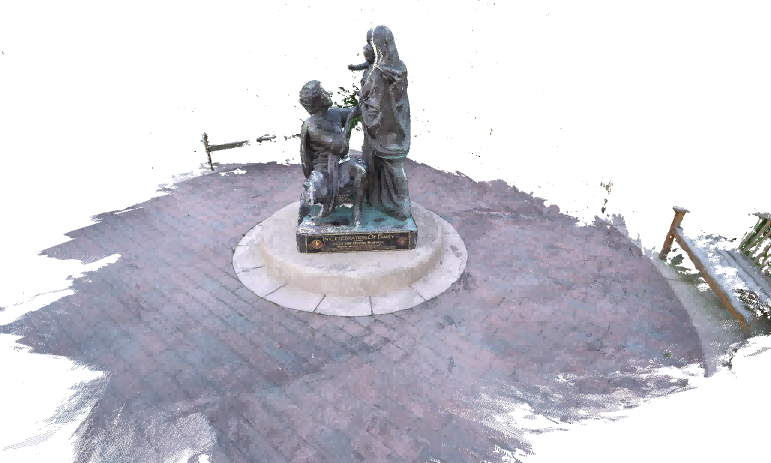}
		\includegraphics[width=.3\linewidth]{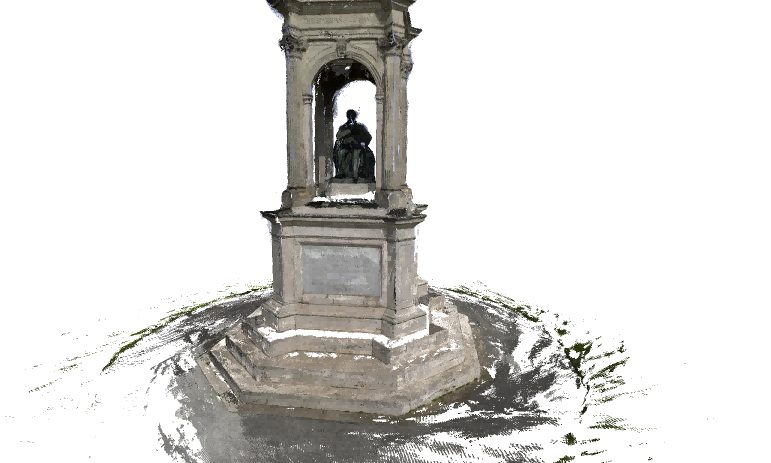}
		\includegraphics[width=.3\linewidth]{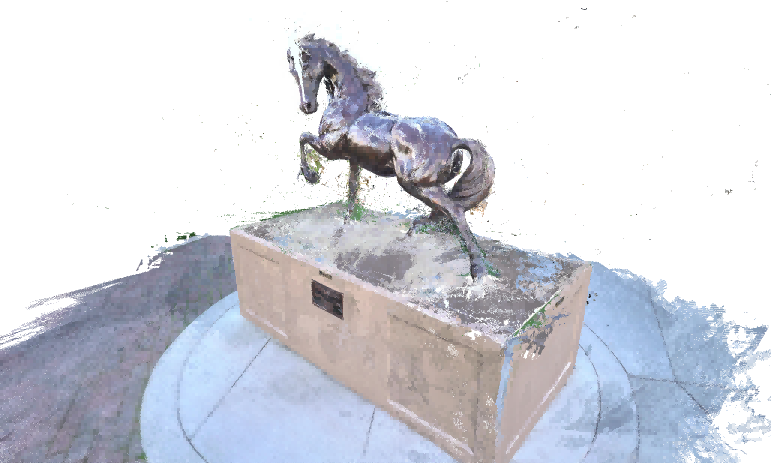}
		
		\includegraphics[width=.3\linewidth]{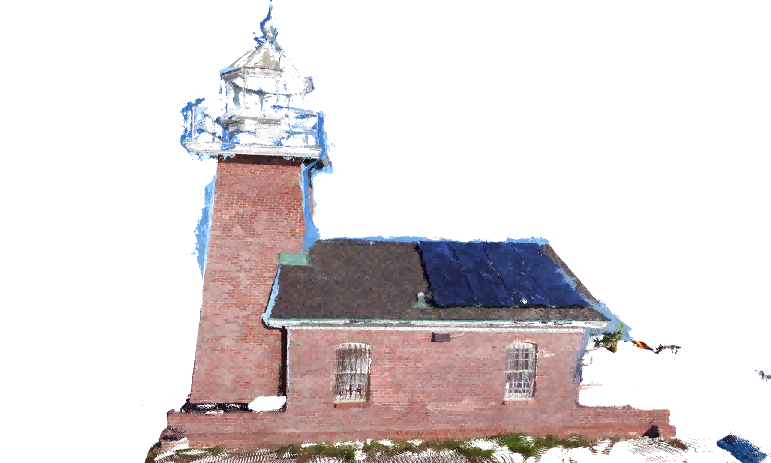}
		\includegraphics[width=.3\linewidth]{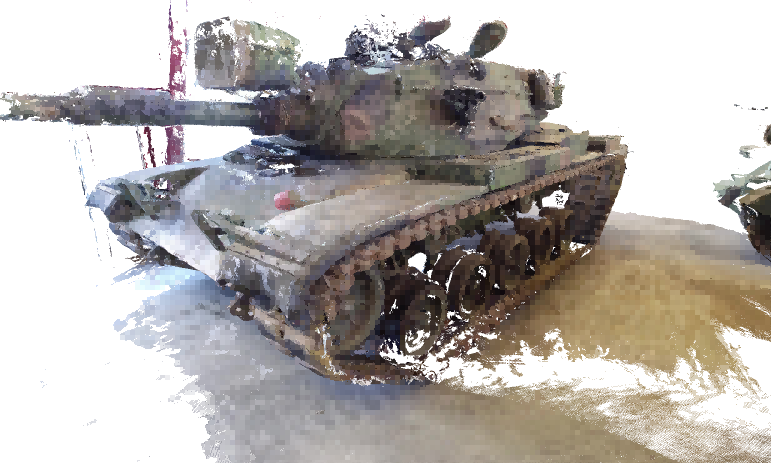}
		\includegraphics[width=.3\linewidth]{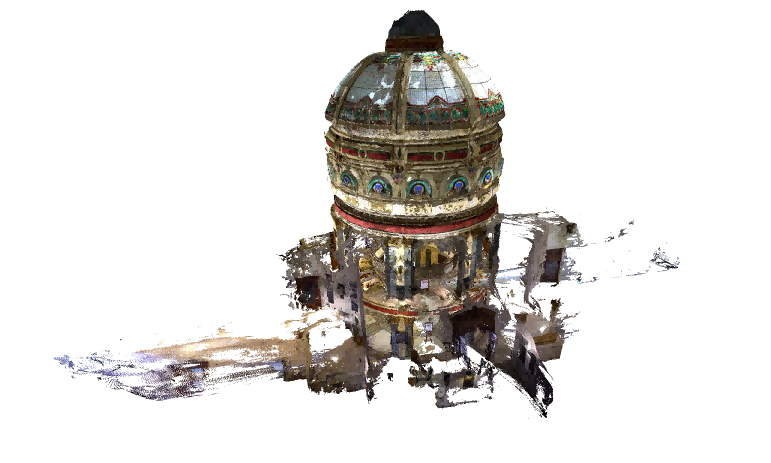}
		
		\includegraphics[width=.3\linewidth]{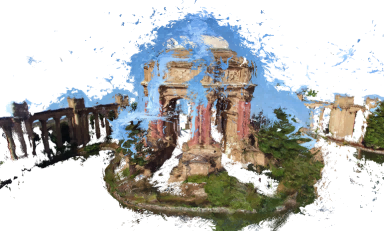}
		\includegraphics[width=.3\linewidth]{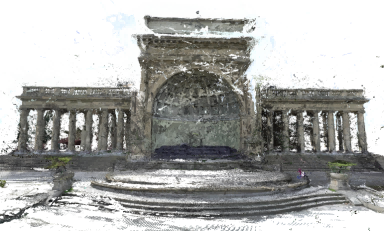}
		\includegraphics[width=.3\linewidth]{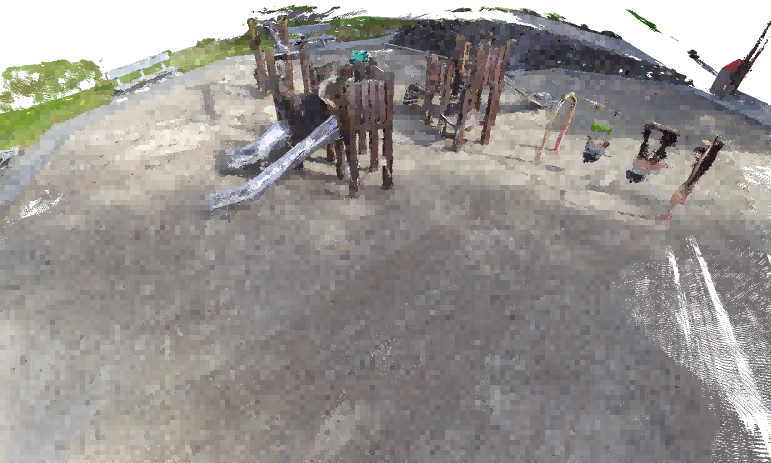}
		
		\includegraphics[width=.45\linewidth]{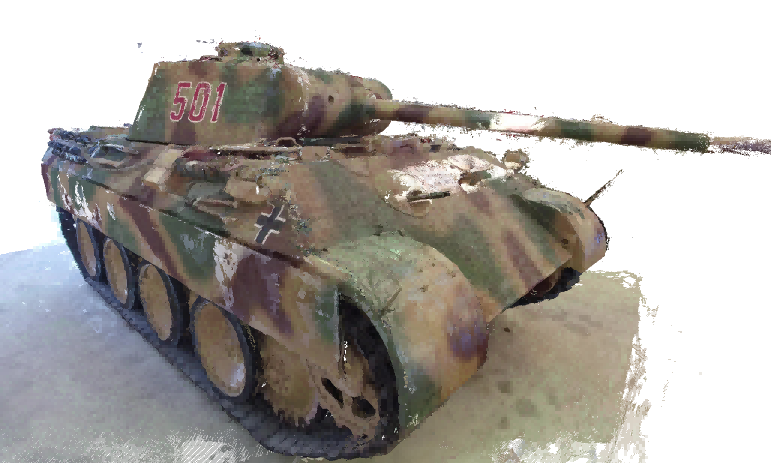}
		\includegraphics[width=.45\linewidth]{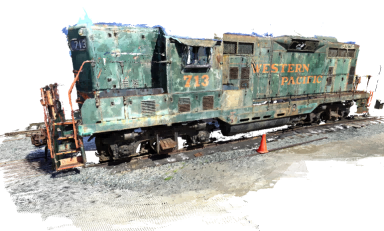}
		
		\caption{
			\textbf{Results on Tanks\&Temple} We visualize all scenes in Intermedia and Advanced.
		}
		
		\label{fig:tanks_pointcloud}
	\end{figure*}

	DispMVS only predicts the depth map and needs to fuse them to point clouds for further evaluation. The fusion can filter the depth map to remove inconsistent regions and reduce the noise in the point cloud. Following MVSNet \cite{yao2018mvsnet}, we use geometric consistency to filger out inconsistent regions and reduce the noise in the point cloud. Geometric consistency contains two steps, the first step projects depth maps of source images to the reference image to mask out regions with large depth differences, and the second step reprojects the depth map of the reference image to source images to mask out regions with large pixel displacements. Besides, DispMVS also uses the depth range to filter out points out of bound. After filtering out these inconsistent regions and invalid regions, DispMVS reproject pixels to 3D space to generate point clouds. 
	Fig. \ref{fig:dtu_pointcloud} and Fig. \ref{fig:tanks_pointcloud} show point clouds generated by DispMVS.
	
	
	\bibliography{aaai23}

\end{document}